\documentclass[11pt]{article}

\usepackage[colormark, titleright, nolineno]{showlab}
\shorttitle{RUIBench}

\usepackage{xspace}
\usepackage{makecell}
\usepackage{nicematrix}
\usepackage{wrapfig}
\usepackage[capitalize]{cleveref}
\crefname{section}{Sec.}{Secs.}
\Crefname{section}{Section}{Sections}
\crefname{table}{Tab.}{Tabs.}
\Crefname{table}{Table}{Tables}
\crefname{appendix}{App.}{Apps.}
\Crefname{appendix}{Appendix}{Appendices}
\crefname{subappendix}{App.}{Apps.}
\Crefname{subappendix}{Appendix}{Appendices}

\definecolor{slGain}{HTML}{3F8F3A}

\newcommand{\RUIBench}{RUIBench\xspace}
\newcommand{\RUIKit}{RUIKit\xspace}
\newcommand{\cmark}{\textcolor{slGain}{\checkmark}}
\newcommand{\xmark}{\textcolor{slGray}{$\times$}}
\newcommand{\pmark}{\textcolor{slOrange!80!black}{$\triangle$}}
\newcommand{\tool}[1]{\texttt{#1}}

\logo{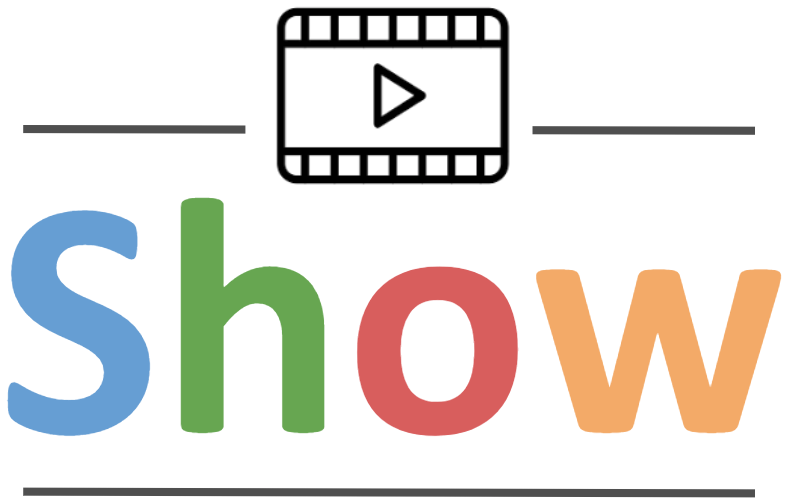}
\headlogoheight{6.5mm}
\runningmark{November, 2026}
\title{\fontsize{16}{20}\selectfont\raisebox{-0.2em}{\includegraphics[height=1.3em]{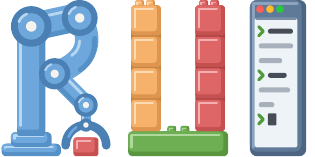}}\hspace{0.3em}RUIBench: What Makes a Good Robot-Use Interface?}

\author{
  Yanzhe Chen\amark[*]{1}\quad
  Zhijun Cao\amark[*]{1}\quad
  Zechen Bai\amark[*]{1}\quad
  Wenzheng Zeng\amark[*]{1}\quad
  Kevin Yuchen Ma\amark{1}\quad
  Yiqi Lin\amark{1}\quad \\
  Guoqiang Liang\amark{1}\quad
  Xiaokang Liu\amark{1}\quad
  Ke Sun\amark{1}\quad
  Kevin Qinghong Lin\quad
  Mike Zheng Shou\amark[\textdagger]{1}
}
\affiliation{Show Lab, National University of Singapore}
\authornote{\textsuperscript{*}Equal contribution.\quad
            \textsuperscript{\textdagger}Corresponding author}
\links{\resource{\textbf{Website [Benchmark \& Code \& Model]} :}{https://showlab.github.io/RUIBench}}

\begin{document}
\maketitle
\thispagestyle{firstpage}

\begin{abstract}
Foundation models can reason about complex physical tasks, yet translating this intelligence into reliable robot execution remains challenging.
We argue that a critical, often overlooked factor is the \textbf{robot-use interface (RUI)}, which determines how agents express and execute physical actions.
To systematically study its impact, we introduce \textbf{RUIBench}, comprising 30 manipulation tasks across five suites spanning world knowledge, in-context learning, discovery, dynamic interaction, and contact-rich manipulation.
We further develop \textbf{RUIKit}, a unified framework that crosses six action abstractions with two interaction forms, yielding 13 interfaces under shared perception and control.
In simulation, changing only the RUI shifts a frontier agent's success rate from \textbf{45\% to 81\%}, and that of a fine-tuned 4B model from \textbf{13\% to 47\%}. Our analysis reveals that effective interfaces \textbf{reduce unnecessary decisions while preserving essential feedback}, with their benefits shaped by agent behavior and physical uncertainty. We further uncover a \textbf{verification gap}, where agents frequently declare success despite observable failures, revealing the need for task-level verification beyond individual actions.
Together, these findings establish RUI design as a critical factor in embodied agent reliability.

\end{abstract}

\begin{figure}[h]
  \centering
  \includegraphics[width=0.98\linewidth]{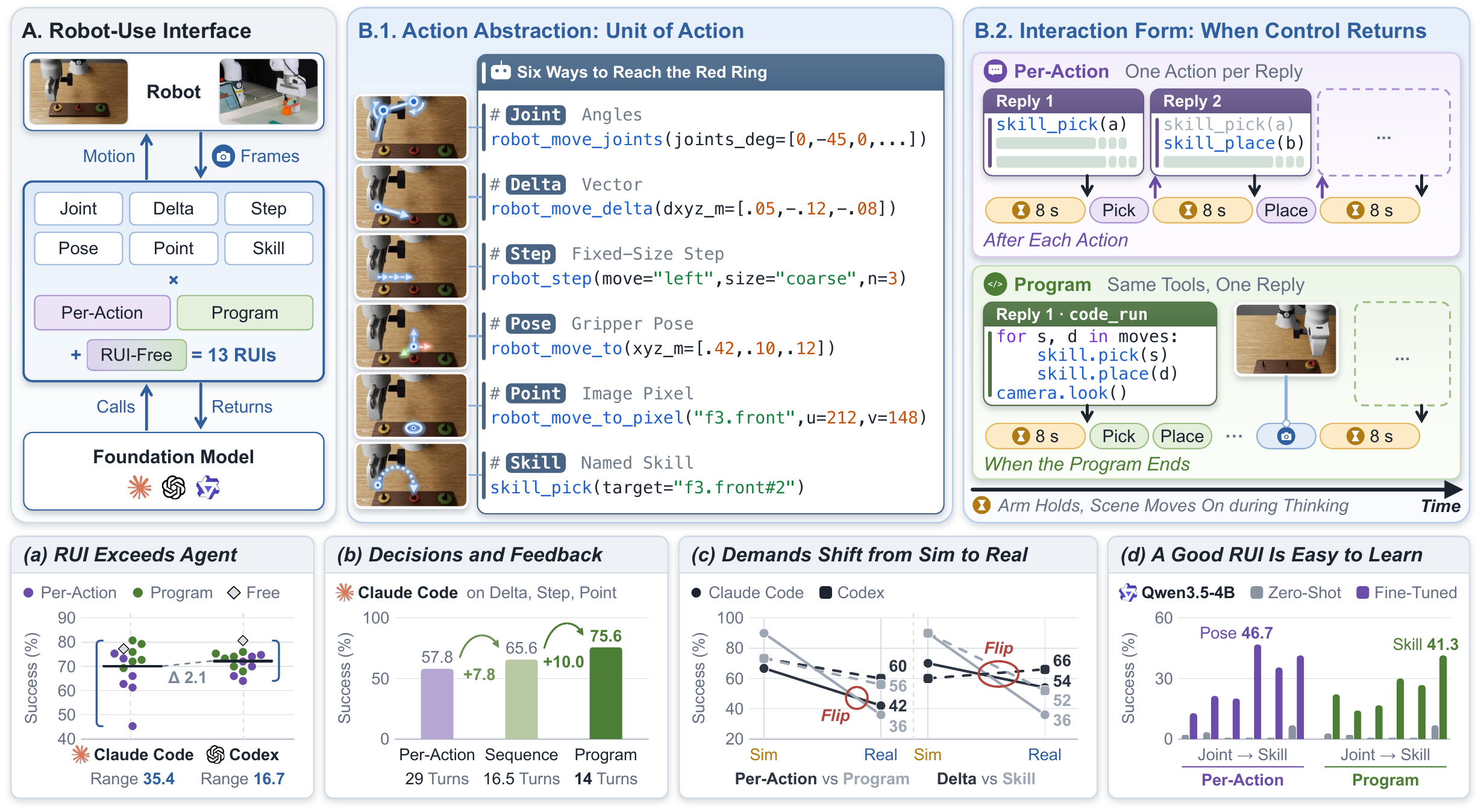}
  \caption{
  \textbf{Robot-use interfaces (RUIs) shape embodied agent performance.}
  \textbf{Top:} \RUIKit organizes six action abstractions and two interaction forms into 13 RUIs under shared perception and control.
  \textbf{Bottom:} (a) Interface choice substantially affects agent success.
  (b) Decision compression and in-program feedback provide complementary gains.
  (c) Interface preferences shift from simulation to reality.
  (d) RUI-specific fine-tuning improves small-model performance in both interaction forms.
  }
  \label{fig:teaser}
  \vspace{-6pt}
\end{figure}

\section{Introduction}
\label{sec:intro}
Foundation models are moving from understanding the physical world to acting in it~\cite{team2025gemini,azzolini2025cosmos,ji2025robobrain}.
While vision-language-action models learn to predict robot actions~\cite{brohan2023rt,kim2024openvla,black2024pi_0}, recent agentic systems enable general-purpose models to control robots through tool calls and executable programs~\cite{liang2023code,vemprala2024chatgpt,chen2026show,fu2026cap,lee2026robotuse,chen2026multimodal}.
This introduces a critical design choice beyond the model itself: how should an agent interact with a robot?
We define the \textbf{robot-use interface (RUI)} as the actions an agent can command and how they are organized for execution.
From low-level motion commands~\cite{belkhale2024rth,berman2026claude} and image-grounded points~\cite{liu2024moka,nasiriany2024pivot} to semantic skills~\cite{ahn2022can,huang2022inner} and executable programs~\cite{liang2023code,singh2022progprompt}, the RUI determines what the agent must decide, what can be delegated to execution, and when control returns to the agent.

This distinction becomes critical in complex physical tasks.
These tasks require agents to gather missing information, react to dynamic events, and correct errors through physical feedback.
An interface that demands a new model decision for every small movement can make execution costly and error-prone, while one that provides insufficient feedback may fail when the world deviates from the plan.
Thus, reliable execution depends not only on what an agent knows, but also on \textbf{how decisions and feedback~are distributed} between the agent and the robot.

However, existing embodied benchmarks compare models under fixed interfaces~\cite{li2024embodied,yang2025embodiedbench,cheng2025embodiedeval,yang2026robotworld,espinola2026reasoning}, while interface studies typically examine selected design choices~\cite{fu2026cap,santos2026alrm,si2026fewer}.
Consequently, the impact of interface design remains poorly understood.
In our experiments, changing only the RUI shifts Claude Code's success rate from 45\% to 81\%, far exceeding the average performance gap between two frontier agents (\cref{fig:teaser}a).
This raises our central question: \textbf{What makes a good robot-use interface?}

To answer this question, we introduce \RUIBench, a benchmark combining 30 agentic manipulation tasks with a controlled evaluation protocol.
These tasks go beyond simple pick-and-place, requiring agents to reason, acquire information through interaction, and adapt to changing physical conditions.
They span five suites: world knowledge, in-context learning, discovery, dynamic interaction, and contact-rich manipulation.
The protocol limits agents to RGB-D observations and robot states, keeps the environment evolving during model reasoning, and evaluates task success independently of the agents' claims.
We further develop \RUIKit, a unified framework that disentangles two design choices: \textbf{action abstraction}, which determines what an agent can command, and \textbf{interaction form}, which determines how actions are organized and executed.
\RUIKit crosses six action abstractions with two interaction forms to instantiate 12 interfaces, plus a free-choice interface, all within a~shared~harness~where~only~the~RUI~varies~(\cref{fig:teaser}).

Across frontier and small models in simulation and 650 real-robot episodes with frontier agents, we find that effective RUIs \textbf{reduce unnecessary decisions while preserving essential feedback} (\cref{fig:teaser}b), which also makes them easier for small models to learn (\cref{fig:teaser}d).
Fine-grained interaction can fragment a clear plan into excessive model decisions, a failure we call \textbf{execution fragmentation}.
Composing actions into programs reduces this burden when agents effectively exploit multi-step execution and intermediate feedback.
However, real-world execution changes which interfaces are effective, as physical uncertainty demands more opportunities for local correction (\cref{fig:teaser}c).
We further uncover a \textbf{verification gap}: agents frequently declare success despite observable failures, while checking individual actions does not guarantee task completion.
Together, these findings highlight the importance of allocating decisions, feedback, and verification at appropriate levels.

Our main contributions are summarized as follows:
\begin{itemize}
    \item \textbf{Agentic benchmark.} We introduce \RUIBench with 30 manipulation tasks across five suites and a controlled evaluation protocol for systematically studying RUIs.
    \item \textbf{Unified framework.} We develop \RUIKit, spanning 13 interfaces across six action abstractions and two interaction forms under shared perception and control.
    \item \textbf{Empirical insights.} We reveal how decision burden, physical feedback, and task-level verification shape embodied performance across agents and environments.
\end{itemize}

\section{Related Work}
\label{sec:related}

\begin{figure}[t!]
  \centering
  \includegraphics[width=\linewidth]{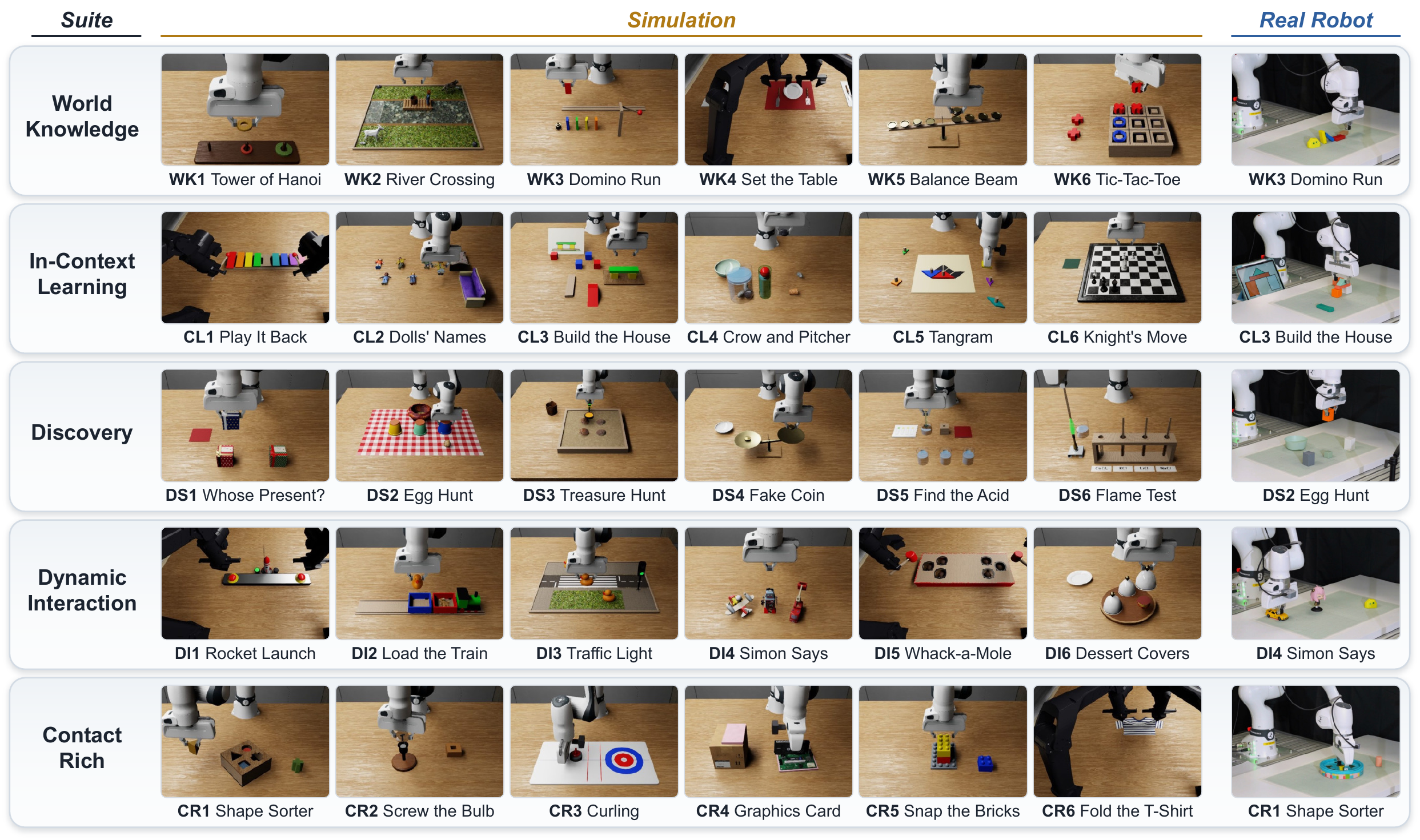}
  \caption{\textbf{The 30 tasks of \RUIBench.}
  Each row is a suite of six simulated tasks, defined by what success requires: reasoning from prior knowledge, using previously shown visual examples, physical exploration, responding to time-sensitive events, or precise adjustments during contact.
  The right column shows~the~task~of~each~suite~that~also~runs~on~the~real~robot.
  }
  \label{fig:tasks}
\end{figure}

\begin{wraptable}{R}{0.535\linewidth}
  \vspace{-16pt}
    \caption{Comparison with related work.
    \textbf{Crossed design} evaluates matched action abstractions under per-action and program execution. \textbf{False claims} evaluates completion claims against physical outcomes. \textbf{Ours: RUIBench +~RUIKit.} \pmark: partial coverage.}
    \label{tab:related}
    \centering
    \footnotesize
    \setlength{\tabcolsep}{0.3pt}
    \renewcommand{\xmark}{\textcolor{slInk}{$\times$}}
    \newcommand{\hd}[1]{{\scriptsize\makecell{#1}}}
    \begin{NiceTabular*}{\linewidth}{@{\extracolsep{\fill}}l*{6}{w{c}{27.2pt}}@{}}[cell-space-limits=1.5pt]
      \CodeBefore
        \rowcolor{black!4}{3,9}
        \rowcolor{slGreen!12}{12}
      \Body
      \toprule
      \Block[l]{2-1}{\hd{Work}}
        & \Block{1-3}{\hd{Interface design}} & &
        & \Block{1-3}{\hd{Evaluation}} & & \\
      \cmidrule(lr){2-4} \cmidrule(lr){5-7}
        & \hd{Action\\abstr.}
        & \hd{Call vs.\\program}
        & \hd{Crossed\\design}
        & \hd{False\\claims}
        & \hd{Small\\models}
        & \hd{Real\\robot} \\
      \midrule
      \Block[l]{1-7}{\emph{Interface studies and systems}} \\
      CaP-X~\cite{fu2026cap}
        & \cmark & \xmark & \xmark
        & \xmark & \cmark & \cmark \\
      ALRM~\cite{santos2026alrm}
        & \xmark & \cmark & \xmark
        & \xmark & \cmark & \xmark \\
      Berman \etal~\cite{berman2026claude}
        & \cmark & \cmark & \xmark
        & \xmark & \xmark & \pmark \\
      Show-Harness~\cite{chen2026show}
        & \cmark & \xmark & \xmark
        & \xmark & \cmark & \cmark \\
      RobotUse~\cite{lee2026robotuse}
        & \pmark & \pmark & \xmark
        & \xmark & \xmark & \cmark \\
      \Block[l]{1-7}{\emph{Agent benchmarks}} \\
      Espinola \etal~\cite{espinola2026reasoning}
        & \xmark & \xmark & \xmark
        & \pmark & \xmark & \xmark \\
      RobotWorld~\cite{yang2026robotworld}
        & \xmark & \xmark & \xmark
        & \pmark & \xmark & \xmark \\
      \midrule
      \textbf{Ours}
    & \cmark & \cmark & \cmark
    & \cmark & \cmark & \cmark \\
      \bottomrule
    \end{NiceTabular*}
  \vspace{-16pt}
\end{wraptable}

\paragraph{Robot-Use Interfaces for Foundation Models}
Foundation models control robots through learned action policies~\cite{brohan2023rt,kim2024openvla,black2024pi_0} or interfaces that connect general-purpose models to physical actions.
Existing systems employ semantic skills and learned-policy tools~\cite{ahn2022can,huang2022inner,rana2023sayplan,yuan2025being,li2026roboclaw,li2026robo,shi2025hi,yang2025agentic,zhang2026harness,chen2026volo,peng2026cortex,galanti2026addressing}, executable programs~\cite{liang2023code,singh2022progprompt,vemprala2024chatgpt,huang2023instruct2act,chen2025code,jia2026agent,wang2026harnesspai}, spatial constraints and value maps~\cite{huang2023voxposer,huang2024rekep,pan2025omnimanip}, image-grounded points and paths~\cite{liu2024moka,nasiriany2024pivot,yuan2024robopoint,team2025gemini,li2025hamster,lee2026robotuse,hu2026via,zhang2026world}, end-effector poses~\cite{kwon2024language}, and discrete motion commands~\cite{belkhale2024rth,chen2026show,guo2026transferring}.
However, these interfaces are typically developed within individual systems, making their independent effects difficult to isolate.
Recent studies compare selected action abstractions~\cite{fu2026cap}, programmatic versus per-action execution~\cite{santos2026alrm,si2026fewer,cho2026spatialclaw}, interface families~\cite{liu2026guava,sensenova2026has,berman2026claude,zhou2026embodied}, harness components~\cite{wang2026justrua,chen2026controlled,fan2026video,sun2026forceharness}, and planner-policy interfaces~\cite{hu2026matters}.
Yet how these design choices jointly shape agent behavior remains poorly understood.
\RUIKit addresses this gap by disentangling \emph{action abstraction} and \emph{interaction form}, enabling controlled comparisons of their individual and combined effects on agent behavior.

\paragraph{Benchmarks for Embodied Foundation Models}
Existing embodied benchmarks evaluate robot manipulation, generalization, and embodied reasoning~\cite{li2024embodied,yang2025embodiedbench,cheng2025embodiedeval,li2026humanclaw}.
Recent efforts extend toward foundation-model agents, introducing tasks that demand multi-step reasoning, planning, and physical interaction~\cite{espinola2026reasoning,yang2026robotworld,shen2026embodiedswe,gao2026hexashell,leet2026bitter,wang2026embodied,gao2026embodiedlab,chen2025code2video}.
Evaluation protocols have also become more realistic, incorporating non-privileged observations, continuous physical time, and independent outcome scoring~\cite{yang2026robotworld,espinola2026reasoning,sensenova2026has}, alongside time and cost considerations~\cite{froger2025are,zhang2025videogamebench,wen2025realtime,kapoor2024ai,kapoor2025holistic}.
However, these benchmarks rarely vary the interface in a controlled way, limiting our understanding of how interface design affects agents facing different interaction demands.
Moreover, physical task success does not necessarily imply that an agent correctly recognizes completion.
Prior studies investigate visual success detection and failure reasoning~\cite{du2023vision,liu2023reflect,duan2024aha,zhou2025code}, yet agents can still declare success despite incomplete tasks~\cite{navasardyan2026failbench,yang2026robotworld,espinola2026reasoning}.
\RUIBench complements these efforts with 30 agentic tasks spanning five interaction demands and a protocol that separates physical outcomes from agents' completion claims.
Together, \RUIBench and \RUIKit measure how interface design \mbox{shapes both execution and completion judgment (\cref{tab:related}).}

\section{RUIBench}
\label{sec:ruibench}

\RUIBench provides the tasks (\cref{fig:tasks}) and evaluation protocol (\cref{sec:protocol}) for studying how RUIs shape embodied agent performance.
For a given agent and task, it fixes the robot, perception, timing, and scoring, and varies the RUI through \RUIKit (\cref{sec:ruikit}).

\subsection{Agentic Task Design}
\label{sec:tasks}

\newlength{\tabtwoep}
\begin{table}[t]
  \caption{\textbf{Statistics of \RUIBench.}
  Source columns count tasks adapted from psychology (Psych.) or drawn from everyday games, puzzles, school experiments, and chores (Everyday).
  Ability columns count tasks emphasizing long-horizon (LH), reasoning (RS), memory (ME), fine-grained control (FG), speed (SP), and bimanual coordination (BI).
  Task statistics are per-task means of instruction length (Instr., words), scene objects, milestones, and scripted-policy time (Ref., s).
  Each task provides 15 development and 20 test missions.}
  \label{tab:tasks}
  \centering
  \footnotesize
  \setlength{\tabcolsep}{2pt}
  \setlength{\tabtwoep}{\dimexpr\textwidth-398pt\relax}
  \begin{NiceTabular}{@{}w{l}{92pt}w{c}{28pt}w{c}{38pt}!{\hspace{4pt}}I*{6}{w{c}{15.5pt}}I>{\hspace{4pt}}w{c}{24pt}w{c}{31pt}w{c}{\tabtwoep}w{c}{32pt}@{}}[cell-space-limits=1pt, custom-line={letter=I, color=black!25}]
    \CodeBefore
      \cellcolor{slBlue!37}{3-4}
      \cellcolor{slBlue!37}{3-5}
      \cellcolor{slBlue!12}{3-7}
      \cellcolor{slBlue!12}{3-9}
      \cellcolor{slBlue!22}{4-4}
      \cellcolor{slBlue!22}{4-5}
      \cellcolor{slBlue!37}{4-6}
      \cellcolor{slBlue!22}{4-7}
      \cellcolor{slBlue!12}{4-9}
      \cellcolor{slBlue!37}{5-4}
      \cellcolor{slBlue!22}{5-5}
      \cellcolor{slBlue!47}{5-6}
      \cellcolor{slBlue!12}{5-7}
      \cellcolor{slBlue!12}{6-6}
      \cellcolor{slBlue!12}{6-7}
      \cellcolor{slBlue!57}{6-8}
      \cellcolor{slBlue!22}{6-9}
      \cellcolor{slBlue!12}{7-4}
      \cellcolor{slBlue!12}{7-5}
      \cellcolor{slBlue!47}{7-7}
      \cellcolor{slBlue!12}{7-9}
      \rowcolor{black!5}{8}
    \Body
    \toprule
    \Block[l]{2-1}{Suite}
      & \Block{1-2}{Source} &
      & \Block{1-6}{Primary abilities} & & & & &
      & \Block{1-4}{Task statistics} & & & \\
    \cmidrule(lr){2-3}
    \cmidrule(lr){4-9}
    \cmidrule(lr){10-13}
      & Psych. & Everyday
      & LH & RS & ME & FG & SP & BI
      & Instr. & Objects & Milestones & Ref. (s) \\
    \midrule
    World knowledge (WK)
      & 3 & 3 & 4 & 4 & 0 & 1 & 0 & 1
      & 16 & 13 & 3.8 & 36 \\
    In-context learning (CL)
      & 2 & 4 & 2 & 2 & 4 & 2 & 0 & 1
      & 14 & 16 & 3.7 & 40 \\
    Discovery (DS)
      & 0 & 6 & 4 & 2 & 5 & 1 & 0 & 0
      & 10 & 10 & 3.5 & 34 \\
    Dynamic interaction (DI)
      & 4 & 2 & 0 & 0 & 1 & 1 & 6 & 2
      & 15 & 9 & 4.0 & 61 \\
    Contact-rich (CR)
      & 2 & 4 & 1 & 1 & 0 & 5 & 0 & 1
      & 14 & 4 & 4.0 & 22 \\
    \midrule
    \textbf{All}
      & 11 & 19 & 11 & 9 & 10 & 10 & 6 & 5
      & 14 & 11 & 3.8 & 38 \\
    \bottomrule
  \end{NiceTabular}
\end{table}

\paragraph{Task design}
A clear task goal does not imply a fully specified path to achieving it.
Successful manipulation often requires agents to reason beyond the initial instruction, acquire information through interaction, and adapt to physical feedback.
\RUIBench captures these demands through 30 tasks in five complementary suites (\cref{fig:tasks}).
\textbf{World knowledge} requires reasoning from prior knowledge, as in the Tower of Hanoi.
\textbf{In-context learning} requires using previously shown visual examples, as in building a house from a picture.
\textbf{Discovery} requires physical exploration, as in finding a fake coin with a balance.
\textbf{Dynamic interaction} requires responding to time-sensitive events, as in Simon Says.
\textbf{Contact-rich manipulation} requires precise adjustments during contact, as in inserting a graphics card.
Negative controls confirm that tasks depend on their intended capabilities (\cref{app:benchmark}).
A scripted reference that solves all 70 control missions succeeds on only 6 of them when it ignores demonstrations, observations after the first frame, or mid-episode changes.
Of the 30 tasks, 11 adapt paradigms from developmental and comparative psychology, and 19 come from everyday games, puzzles, school experiments, and chores (\cref{tab:tasks,tab:app-tasks}).
These familiar settings make goals intuitive and outcomes interpretable while preserving the challenges of agentic decision-making.

\paragraph{Task configurations}
Each suite contains six tasks, with each task drawing on at least three of six underlying abilities: long-horizon planning, reasoning, memory, fine-grained control, speed, and bimanual coordination.
The suites emphasize different combinations of these abilities (\cref{tab:tasks}).
Each task includes three or four milestones to measure partial progress.
Twenty-five tasks use a Franka Research~3 arm with fixed and wrist cameras, while five use two ARX~X5 arms with head and wrist cameras.
All tasks are implemented in Isaac Sim with Isaac Lab~\cite{mittal2025isaac}.
Each task provides 15 development and 20 test missions from fixed seeds.
We retain only layouts that pass reachability, visibility, and graspability checks and can be solved by a scripted reference policy with privileged state, ensuring feasibility in simulation.
Prompts, task parameters, and scorers are configured on development missions, while all agents and RUIs are evaluated on the same held-out test missions.

\subsection{Evaluation Protocol}
\label{sec:protocol}

To isolate interface effects while preserving the demands of physical interaction, \RUIBench enforces non-privileged observations, continuous physical time, and independent outcome evaluation.

\paragraph{Non-privileged observations}
Agents receive task instructions, hints available to human operators, and, for in-context learning tasks, a previously shown video or image.
All other information comes from RGB-D cameras and robot-state readings through shared tools.
SAM~3~\cite{carion2025sam} detects and segments objects, while 3D positions are recovered from noisy sensor depth rather than privileged simulator states.
On 100 annotated frames, this pipeline detects 93\% of objects with localization errors of 0.5 to 0.7\,cm.\looseness=-1

\paragraph{Continuous physical time}
The environment continues evolving while the agent reasons.
In simulation, each model response incurs a fixed 8\,s delay, during which the robot holds its pose while other objects continue moving.
This approximates the 75th percentile of 4,067 responses measured from three frontier models, preventing differences in serving latency from confounding interface comparisons.

\paragraph{Independent outcome evaluation}
Each episode starts from a fresh simulator instance and ends when the agent declares completion, gives up, or reaches its turn budget or 30\,min time limit.
For reference, scripted policies complete tasks in 38\,s on average (\cref{tab:tasks}), while some dynamic events last only 16 to 20\,s.
Reaching the goal does not end an episode.
Instead, success is determined independently from the physical trajectory, verified offline from execution logs, and withheld from the agent.
The agent must therefore decide when the task is complete, allowing its completion claims~to~be~compared~against~actual~physical~outcomes.

\paragraph{Evaluation measures}
Beyond task success and API cost per episode, we measure two aspects of interface behavior.
\textbf{Decision burden} is characterized by model turns per episode and, for programs, actions executed per program.
\textbf{Completion judgment} is evaluated through three rates (\cref{tab:judgment}): the fraction of episodes declared complete, the fraction of completion claims that are false, and the fraction of failed episodes that end in a completion claim.
Together, these measures distinguish execution performance from the agent's ability to recognize task completion.

\section{RUIKit}
\label{sec:ruikit}

\begin{wrapfigure}{R}{0.53\linewidth}
  \vspace{-16pt}
  \centering
  \includegraphics[width=\linewidth]{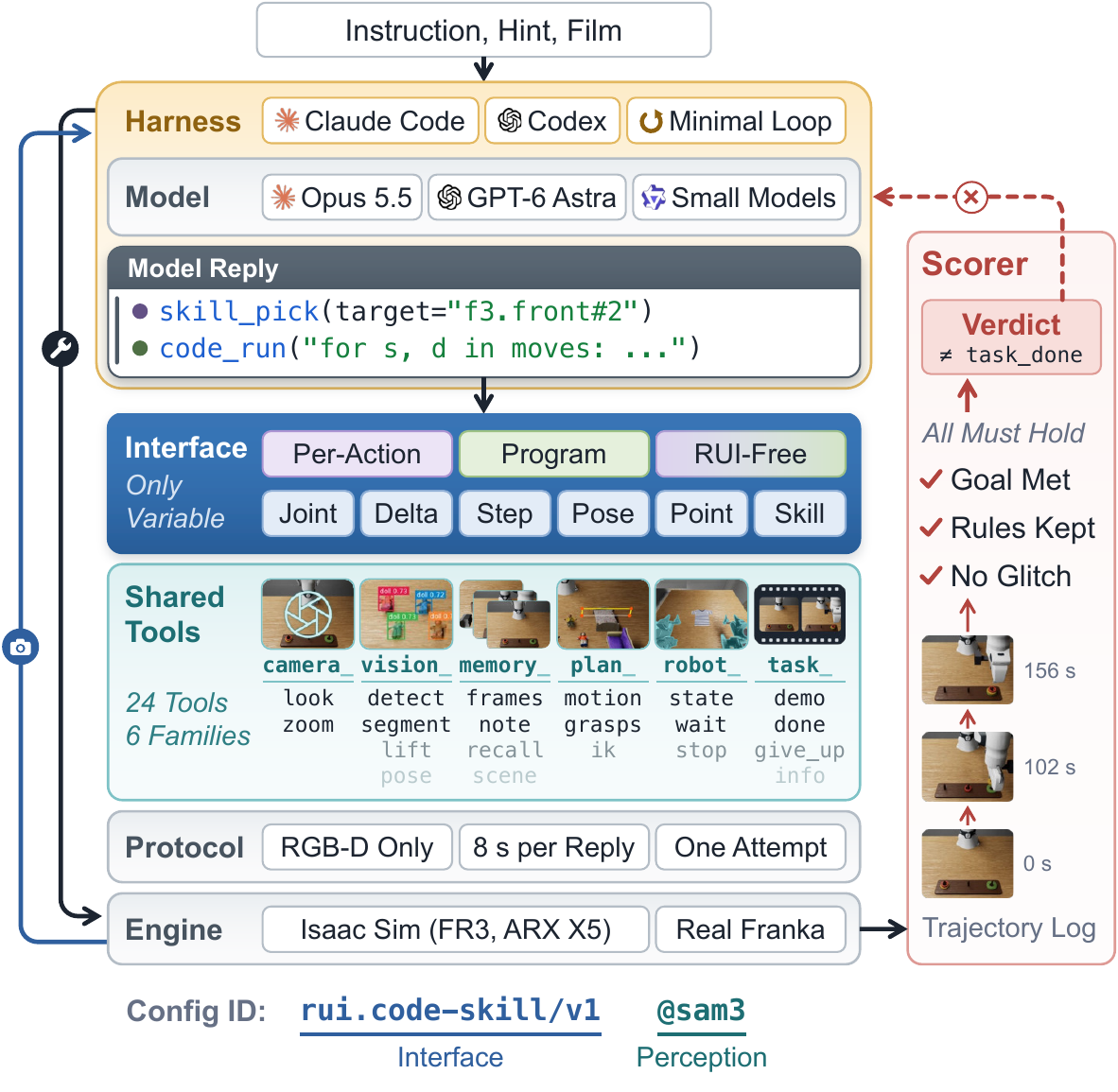}
  \caption{\textbf{Overview of \RUIKit.}
  The RUI is the only variable for a fixed agent and task, while the harness, 24 underlying tools, evaluation protocol, and robot backend remain shared.
  The scorer's verdict is withheld from the agent.\looseness=-1}
  \label{fig:ruikit}
  \vspace{-52pt}
\end{wrapfigure}

\RUIKit provides a unified framework for constructing and evaluating robot-use interfaces (\cref{fig:ruikit}).
It separates interface design from the underlying perception, control, and agent infrastructure.
For a given agent and task, \textbf{only the RUI varies}, while the harness, tools, evaluation protocol, and simulator~remain~fixed~across~RUIs.

\subsection{Interface Design Space}
\label{sec:design-space}

Robot-use interfaces differ in what actions they expose and how those actions are organized for execution.
We characterize this design space along two complementary axes: \textbf{action abstraction} and \textbf{interaction form} (\cref{fig:teaser}).

\paragraph{Action abstraction}
This axis determines what an agent can command.
We consider six abstractions:
Joint specifies joint-space targets;
Delta specifies relative translations and rotations;
Step provides fixed-magnitude motion commands;
Pose specifies absolute end-effector targets;
Point grounds actions in image-space locations; and
Skill exposes semantic operations such as picking, placing, pressing, pushing, and turning.
They differ in representation and \mbox{control granularity rather than forming a strict hierarchy.}

\paragraph{Interaction form}
This axis determines how actions are composed and when control returns to the agent.
In the \emph{per-action} form, each model reply issues at most one action per arm, after which control returns to the agent.
In the \emph{program} form, the agent writes a Python program~\cite{liang2023code} that composes actions through sequences, loops, and conditional branches.
Programs can also query observations and adapt subsequent actions without another model reply.
For a given abstraction, \textbf{both forms use the same action primitives}, so they differ only in how actions are organized.

\paragraph{Interface configurations}
Crossing six action abstractions with two interaction forms yields 12 fixed RUIs.
We also introduce RUI-Free, which exposes all abstractions in both forms and lets the agent choose, giving 13 interfaces.
Per-action RUIs are named by their abstraction, such as \texttt{skill}, while program-based RUIs use the prefix \texttt{program}, such as \texttt{program-skill}.
Importantly, Skill and Program represent different forms of delegation: \emph{skills encapsulate predefined robot operations}, whereas \emph{programs compose operations through agent-generated code}.
They can therefore be combined rather than treated as alternatives.
All configurations share the same infrastructure, allowing~either~axis~to~vary~while~the~other~stays~fixed.

\subsection{Unified Agent Harness}
\label{sec:harness}

\paragraph{Shared perception and control}
\RUIKit provides 24 underlying tools across six families (\cref{fig:ruikit,tab:app-rui}).
All RUIs share the same perception and control infrastructure, including camera observation, object detection and segmentation, depth-based 3D localization, and robot-state estimation.
Shared utilities further support memory, visual marking, object tracking, grasp and motion planning, and reachability checks.
Agents also access task information, such as instructions, demonstrations, and events, and issue waiting, stopping, completion, or give-up commands through common tools.
Only the exposed action commands and their interaction form vary across RUIs.\looseness=-1

\paragraph{Execution and agent integration}
\RUIKit implements robot actions and observations as reusable functions.
Each interface selects its action primitives and interaction form, with a versioned identifier such as \texttt{rui.code-skill/v1}.
Frontier coding agents access these functions through the Model Context Protocol (MCP), while smaller open models use a minimal tool-calling loop.
Programs execute in a sandbox exposing only permitted functions.
Both agent types use \mbox{the same robot-side infrastructure and interface definitions.}

\paragraph{Validation and deployment}
\RUIKit records tool calls, robot trajectories, observations, and outcomes for offline evaluation.
An independent scorer evaluates goal satisfaction, task constraints, and simulation validity from the logged trajectories.
Each session is also audited for unauthorized tool use, file or network operations, and access to benchmark internals.
A real-robot backend for the Franka arm reuses the same interface definitions, tools, and harness, enabling all 13 RUIs to run without modification.

\newlength{\tthreew}
\newlength{\tthreewp}
\begin{table}[t]
  \caption{\textbf{Simulation results across robot-use interfaces.}
  Success rates are reported for five task suites and their average (Avg.).
  Turns: median model replies per episode; Cost: mean API cost per episode (\$).
  For Qwen3.5-4B, ZS/FT: zero-shot/fine-tuned success; Train/Test: one-time fine-tuning cost and cost per test episode, at \$2.68 per H200 GPU-hour.
  Bold: best success and lowest per-episode cost per agent.
  }
  \label{tab:main}
  \centering
  \scriptsize
  \setlength{\tabcolsep}{1pt}
  \setlength{\tthreew}{\dimexpr(\textwidth-92.6pt)/20\relax}
  \setlength{\tthreewp}{\dimexpr\tthreew+2.5pt\relax}
  \begin{NiceTabular}{@{}w{l}{35pt}*{8}{w{c}{\tthreew}}!{\hspace{2.5pt}}I>{\hspace{2.5pt}}w{c}{\tthreewp}*{7}{w{c}{\tthreew}}!{\hspace{2.5pt}}I>{\hspace{2.5pt}}w{c}{\tthreewp}*{2}{w{c}{\tthreew}}w{c}{\tthreewp}<{\hspace{2.5pt}}@{}}[cell-space-limits=1.3pt,custom-line={letter=I,color=black!25}]
    \CodeBefore
      \rowcolor{slBlue!8}{4}
      \rowcolor{slBlue!4}{11}
      \rowcolor{slGreen!10}{12}
      \rowcolor{slGreen!5}{19}
      \rowcolor{black!4}{20}
      \rectanglecolor{black!3}{5-7}{10-7}
      \rectanglecolor{black!3}{13-7}{18-7}
      \rectanglecolor{black!3}{21-7}{22-7}
      \rectanglecolor{black!3}{5-15}{10-15}
      \rectanglecolor{black!3}{13-15}{18-15}
      \rectanglecolor{black!3}{21-15}{22-15}
      \rectanglecolor{black!3}{5-18}{10-19}
      \rectanglecolor{black!3}{13-18}{18-19}
      \rectanglecolor{black!3}{21-18}{22-19}
    \Body
    \toprule
    \Block[l]{3-1}{RUI} & \Block{1-8}{Claude Code} & & & & & & & & \Block{1-8}{Codex} & & & & & & & & \Block{1-4}{Qwen3.5-4B} & & & \\
    \cmidrule(lr){2-9} \cmidrule(lr){10-17} \cmidrule(lr){18-21}
    & \Block{1-6}{Success (\%)} & & & & & & \Block{1-2}{Efficiency} & & \Block{1-6}{Success (\%)} & & & & & & \Block{1-2}{Efficiency} & & \Block{1-2}{Success (\%)} & & \Block{1-2}{Cost (\$)} & \\
    \cmidrule(lr){2-7} \cmidrule(lr){8-9} \cmidrule(lr){10-15} \cmidrule(lr){16-17} \cmidrule(lr){18-19} \cmidrule(lr){20-21}
    & WK & CL & DS & DI & CR & \textbf{Avg.} & Turns & Cost & WK & CL & DS & DI & CR & \textbf{Avg.} & Turns & Cost & ZS & FT & Train & Test \\
    \midrule
    \Block[l]{1-21}{\textbf{Per-action}\quad\emph{one action per arm per model reply}} \\
    Joint & 80 & 77 & 63 & 70 & 40 & 66.0 & 35 & 1.48 & 83 & 77 & 67 & 53 & 70 & 70.0 & 32 & 2.15 & 2.0 & 12.7 & 0.4K & 2.07 \\
    Delta & 93 & 47 & 50 & 70 & 53 & 62.7 & 31 & 1.34 & 80 & 80 & 67 & 37 & 67 & 66.0 & 23 & 1.71 & 3.3 & 21.3 & 0.3K & 1.66 \\
    Step & 67 & 10 & 63 & 40 & 47 & 45.3 & 37 & 2.01 & 87 & 63 & 77 & 67 & 67 & 72.0 & 38 & 2.37 & 0.0 & 20.0 & 0.9K & 0.97 \\
    Pose & 87 & 67 & 80 & 87 & 57 & 75.3 & 29 & 1.19 & 93 & 83 & 60 & 67 & 70 & 74.7 & 23 & 1.58 & 0.0 & \textbf{46.7} & 0.4K & 0.56 \\
    Point & 87 & 50 & 70 & 50 & 50 & 61.3 & 28 & 1.49 & 53 & 67 & 93 & 33 & 73 & 64.0 & 20 & 1.42 & 0.0 & 35.3 & 0.3K & 0.53 \\
    Skill & 100 & 63 & 70 & 63 & 70 & 73.3 & 19 & 0.90 & 87 & 90 & 90 & 57 & 47 & 74.0 & 16 & 1.09 & 6.7 & 41.3 & 0.2K & 0.36 \\
    \textbf{Overall} & 86 & 52 & 66 & 63 & 53 & 64.0 & 30 & 1.40 & 81 & 77 & 76 & 52 & 66 & 70.1 & 25 & 1.72 & 2.0 & 29.6 & 0.4K & 1.03 \\
    \midrule
    \Block[l]{1-21}{\textbf{Program}\quad\emph{one Python program per reply using the same action primitives}} \\
    Joint & 100 & 80 & 67 & 83 & 50 & 76.0 & 12 & 0.81 & 63 & 83 & 87 & 67 & 67 & 73.3 & 27 & 1.75 & 2.7 & 22.0 & 0.2K & 1.98 \\
    Delta & 97 & 70 & 97 & 83 & 57 & \textbf{80.7} & 11 & \textbf{0.69} & 87 & 87 & 63 & 47 & 67 & 70.0 & 20 & 1.74 & 2.0 & 14.0 & 0.2K & 1.96 \\
    Step & 100 & 50 & 93 & 67 & 53 & 72.7 & 14 & 0.77 & 87 & 70 & 80 & 53 & 50 & 68.0 & 34 & 2.27 & 0.0 & 16.7 & 0.3K & 0.55 \\
    Pose & 100 & 70 & 73 & 67 & 50 & 72.0 & 11 & 0.78 & 100 & 83 & 67 & 83 & 47 & 76.0 & 21 & 1.33 & 0.0 & 30.0 & 0.2K & \textbf{0.33} \\
    Point & 100 & 87 & 97 & 80 & 33 & 79.3 & 14 & 0.74 & 87 & 87 & 60 & 87 & 50 & 74.0 & 22 & 1.75 & 0.0 & 26.7 & 0.4K & 0.76 \\
    Skill & 93 & 47 & 67 & 93 & 47 & 69.3 & 12 & 0.86 & 60 & 70 & 93 & 70 & 83 & 75.3 & 14 & \textbf{0.95} & 6.7 & 41.3 & 0.2K & 0.52 \\
    \textbf{Overall} & 98 & 67 & 82 & 79 & 48 & 75.0 & 12 & 0.77 & 81 & 80 & 75 & 68 & 61 & 72.8 & 23 & 1.63 & 1.9 & 25.1 & 0.2K & 1.02 \\
    \midrule
    \Block[l]{1-21}{\textbf{Free choice}\quad\emph{all action abstractions and both interaction forms}} \\
    RUI-Free & 93 & 70 & 80 & 87 & 57 & 77.3 & 14 & 0.78 & 97 & 87 & 67 & 97 & 57 & \textbf{80.7} & 15 & 1.47 & -- & -- & -- & -- \\
    \midrule
    \textbf{All RUIs} & 92 & 61 & 75 & 72 & 51 & 70.1 & 21 & 1.06 & 82 & 79 & 75 & 63 & 63 & 72.2 & 23 & 1.66 & 1.9 & 27.3 & 0.3K & 1.02 \\
    \bottomrule
  \end{NiceTabular}
\end{table}

\section{Experiments}
\label{sec:experiments}

\subsection{Experimental Setup}
\label{sec:setup}

\paragraph{Models}
We evaluate two frontier coding agents, Claude Code with Claude Opus~5.5 and Codex with GPT-6 Astra, both at high reasoning effort and connected through MCP (\cref{sec:harness}).
We also evaluate \mbox{Qwen3.5-4B}~\cite{qwen3.5} through a minimal agent loop without thinking, both zero-shot and after fine-tuning.
For each of the 12 fixed RUIs, we fine-tune a LoRA adapter~\cite{hu2022lora} on 300 training trajectories, 30\% from successful Qwen3.8-Max episodes and 70\% from scripted policies.
About half of the scripted trajectories include injected deviations and recovery.
Training and test seeds are disjoint (\cref{app:qwen}).

\paragraph{Protocol and metrics}
Each frontier agent runs all 30 tasks with 13 RUIs on five test missions per task, yielding 1,950 simulation episodes per agent.
The small model is evaluated on the 12 fixed RUIs, as RUI-Free has no dedicated training data.
All experiments follow the protocol in \cref{sec:protocol}.
We report task success, median model turns per episode, and mean API cost.
For the small model, we also report the one-time fine-tuning cost and the per-episode inference cost at GPU rental prices (Train and Test in \cref{tab:main}).\looseness=-1

\subsection{Interface Effects in Simulation}
\label{sec:sim}

\paragraph{Finding 1: Embodied performance depends on both the agent and its interface}
Changing only the RUI shifts Claude Code's success from 45.3\% to 80.7\%, far exceeding the 2.1-point difference between the frontier agents averaged across all RUIs (\cref{tab:main}).
API cost also varies by nearly threefold.
This shows that the interface is not merely a means of issuing actions: it determines how many decisions an agent makes and what each decision must accomplish.
Importantly, the same interface does not benefit every agent equally.
For Claude Code, programs raise average success from 64.0\% to 75.0\%, reducing median turns from 30 to 12 and mean API cost from \$1.40 to \$0.77.
Claude Code composes 3.3 actions per program, and 54\% of its programs observe the world after acting.
Codex averages 0.9 actions and typically wraps a single action in code, gaining only 2.7 points from the program form (\cref{app:programs}).
Skills offer another way to reduce model decisions by encapsulating predefined operations, achieving the fewest turns and lowest costs among per-action RUIs for both agents.
RUI-Free further illustrates the role of interface selection: it achieves 80.7\% for Codex, outperforming every fixed RUI, while remaining competitive for Claude Code.
Together, these results suggest that \textbf{interface affordances matter only when agents can effectively use them}.

\paragraph{Learning to use an interface}
The small-model experiments examine whether this ability can be acquired through fine-tuning rather than assumed from model scale.
Zero-shot, \mbox{Qwen3.5-4B} succeeds in only 1.9\% of episodes across the 12 fixed RUIs, and does best with Skill (6.7\%), which encapsulates whole operations.
Fine-tuning on 300 trajectories per RUI raises its success to 27.3\% at a one-time cost of \$0.2K to \$0.9K, but how much it learns depends on the RUI (\cref{tab:main}).
Abstractions that name a target rather than a motion (Pose, Point, and Skill) reach 35.3\% to 46.7\% in the per-action form, whereas Joint, Delta, and Step stay between 12.7\% and 21.3\%.
They also shorten episodes: each costs \$0.36 to \$0.56, less than any frontier configuration, whereas a Joint or Delta episode costs \$1.66 to \$2.07.
Programs, in contrast, lower the average success from 29.6\% to 25.1\% and help only with Joint, whose per-action episodes run longest.
Thus, \textbf{a good RUI is also easy to learn}: the small model learns most where the abstraction removes decisions and per-action execution keeps feedback.
Pose and Skill are the two best per-action RUIs for all three agents, even though Pose scores 0.0\% zero-shot.
With per-action Pose, the small model even matches Claude Code with per-action Step (46.7\% vs.\ 45.3\%).

\subsection{Decision Compression and Feedback}
\label{sec:ablation}

\begin{wraptable}{r}{0.53\linewidth}
  \vspace{-16pt}
  \caption{\textbf{Effects of interaction form and action-level checks.}
  Success (\%); Turns: median model replies per episode; Cost: mean API cost per episode (\$); Failed claimed: the share of failed episodes that end in a completion claim (\%). One seed per task.}
  \label{tab:ablation}
  \centering
  \footnotesize
  \setlength{\tabcolsep}{2pt}
  \begin{NiceTabular*}{\linewidth}{@{\extracolsep{\fill}}l*{4}{w{c}{36pt}}@{}}[cell-space-limits=1pt]
    \CodeBefore
      \rowcolor{slBlue!6}{4}
      \rowcolor{slGreen!4}{5}
      \rowcolor{slGreen!8}{6}
    \Body
    \toprule
    \Block[l]{2-1}{Variant} & \Block{2-1}{Success} & \Block{1-2}{Efficiency} & & \Block{2-1}{Failed\\claimed} \\
    \cmidrule(lr){3-4}
    & & Turns & Cost & \\
    \midrule
    \Block[l]{1-5}{\textbf{Interaction form}\enspace\emph{Delta, Step, Point; 90 episodes each}} \\
    Per-action         & 57.8 & 29   & 1.58 & 71.1 \\
    Action sequence    & 65.6 & 16.5 & 0.86 & 77.4 \\
    Program            & \textbf{75.6} & 14 & 0.79 & 86.4 \\
    \midrule
    \Block[l]{1-5}{\textbf{Action-level checks}\enspace\emph{Skill, Program-Skill; 60 episodes each}} \\
    No check           & 73.3 & 16   & 0.89 & 87.5 \\
    Report             & 73.3 & 17.5 & 0.94 & 81.2 \\
    Report and correct & 66.7 & 16.5 & 0.91 & \textbf{75.0} \\
    \bottomrule
  \end{NiceTabular*}
  \vspace{-14pt}
\end{wraptable}

\paragraph{Finding 2: Efficient execution requires both fewer decisions and effective feedback}
Fine-grained per-action control can fragment a clear plan into repeated model decisions, increasing latency, cost, and opportunities for error.
We call this failure \textbf{execution fragmentation}.
Programs can reduce this burden, but their benefits may come from both action composition and intermediate feedback.

To distinguish these effects, we introduce an \emph{action sequence} variant that executes multiple actions without intermediate observations and returns their results only at the end.
On the Delta, Step, and Point interfaces, Claude Code improves from 57.8\% with per-action execution to 65.6\% with action sequences, while median turns fall from 29 to 16.5 (\cref{tab:ablation}).
Allowing observations and conditional execution within programs further raises success to 75.6\%, with 14 turns.
Thus, action composition reduces decision burden, and in-program feedback adds further gains.\looseness=-1

These benefits depend on task demands (\cref{tab:app-suites}).
In-context learning tasks benefit substantially from executing known action sequences, while knowledge, discovery, and dynamic tasks rely more on intermediate feedback.
Contact-rich manipulation benefits from neither consistently, suggesting that its required corrections may exceed what current program-level interaction can provide.
The key distinction is therefore not simply short versus long \mbox{execution, but \textbf{where feedback can influence the next action}}.

\subsection{From Simulation to Reality}
\label{sec:real}

\paragraph{Real-robot setup}
We evaluate both frontier agents across all 13 RUIs on a Franka Research~3 arm with fixed and wrist cameras.
The evaluation covers five tasks, one per suite: Domino Run (WK3), Build the House (CL3), Egg Hunt (DS2), Simon Says (DI4), and Shape Sorter (CR1).
With five layouts per task, this yields 650 real-robot episodes.
Interfaces and task budgets match simulation, while execution follows wall-clock time.
A human operator independently assesses task success, with additional quality ratings and rollouts reported in \cref{app:real}.

\begin{wraptable}{r}{0.56\linewidth}
  \vspace{-16pt}
  \caption{\textbf{Sim-to-real effects of interface design.}
  Success (\%) on five matched tasks (Sim: one seed; Real: 650 episodes).
  $\Delta$: Real minus Sim (percentage points).
  Abstractions average both forms; bold marks the best Real result in each group.}
  \label{tab:real}
  \centering
  \footnotesize
  \setlength{\tabcolsep}{1.2pt}
  \newcommand{\dropval}[1]{\textcolor{slGray}{\ensuremath{-#1}}}
  \newcommand{\riseval}[1]{\textcolor{slGain}{\ensuremath{+#1}}}
  \begin{NiceTabular*}{\linewidth}{@{\extracolsep{\fill}}l*{6}{w{c}{29pt}}@{}}[cell-space-limits=1.1pt]
    \CodeBefore
      \rowcolor{slBlue!6}{4}
      \rowcolor{slGreen!8}{5}
    \Body
    \toprule
    \Block[l]{2-1}{RUI} & \Block{1-3}{Claude Code} & & & \Block{1-3}{Codex} & & \\
    \cmidrule(lr){2-4} \cmidrule(lr){5-7}
    & Sim & \textbf{Real} & $\Delta$ & Sim & \textbf{Real} & $\Delta$ \\
    \midrule
    \Block[l]{1-7}{\textbf{Interaction form / choice}} \\
    Per-action & 66.7 & 42.0 & \dropval{24.7} & 73.3 & \textbf{60.0} & \dropval{13.3} \\
    Program & 90.0 & 36.0 & \dropval{54.0} & 73.3 & 56.0 & \dropval{17.3} \\
    RUI-Free & 100.0 & \textbf{44.0} & \dropval{56.0} & 100.0 & 44.0 & \dropval{56.0} \\
    \midrule
    \Block[l]{1-7}{\textbf{Action abstraction}} \\
    Joint & 70.0 & 34.0 & \dropval{36.0} & 60.0 & 54.0 & \dropval{6.0} \\
    Delta & 70.0 & \textbf{54.0} & \dropval{16.0} & 60.0 & \textbf{66.0} & \riseval{6.0} \\
    Step & 60.0 & 36.0 & \dropval{24.0} & 60.0 & 52.0 & \dropval{8.0} \\
    Pose & 80.0 & 42.0 & \dropval{38.0} & 100.0 & 64.0 & \dropval{36.0} \\
    Point & 100.0 & 32.0 & \dropval{68.0} & 70.0 & 60.0 & \dropval{10.0} \\
    Skill & 90.0 & 36.0 & \dropval{54.0} & 90.0 & 52.0 & \dropval{38.0} \\
    \midrule
    \textbf{All RUIs} & 80.0 & 39.4 & \dropval{40.6} & 75.4 & 56.9 & \dropval{18.5} \\
    \bottomrule
  \end{NiceTabular*}
  \vspace{-16pt}
\end{wraptable}

\paragraph{Finding 3: Real-world execution changes interface preferences}
Interface advantages observed in simulation do not necessarily transfer to physical execution (\cref{tab:real}).
For Claude Code, programs outperform per-action execution by 23.3 points in simulation but lose this advantage on the real robot (36.0\% vs.\ 42.0\%).
Their sim-to-real drops are 54.0 and 24.7 points, respectively.
Codex shows a smaller shift, with per-action execution slightly outperforming programs on the real robot.
Free choice transfers worst: RUI-Free drops from 100.0\% in simulation to 44.0\% on the real robot for both agents.
Among action abstractions, Delta leads for both agents, with Pose close behind for Codex.

These differences highlight the importance of \textbf{feedback and local correction under physical uncertainty}.
More than in simulation, real execution is affected by depth noise, calibration offsets, and slipping grasps, with the two cameras sometimes locating the same object up to 1.5\,cm apart.
Successful trajectories often involve placing an object, inspecting the result, and making incremental corrections.
In contrast, a Claude Code program placed three dominoes without checking their spacing and incorrectly declared completion.
Such cases suggest that the benefit of action composition depends not only on reducing model decisions, but also on \textbf{preserving opportunities to correct execution errors}.\looseness=-1

\paragraph{Limits of current interfaces}
Some tasks require feedback at timescales or precision levels beyond what model-level interaction can reliably provide.
On the real robot, Simon Says remains challenging because the target moves while the agent reasons.
Codex's image-point programs provide an instructive exception, using in-program observations to track and reach the target across all five layouts.
This suggests that time-critical reactions may need to operate within programs or lower-level controllers.

Contact-rich manipulation poses a complementary challenge.
In simulation, inserting a graphics card and folding a T-shirt remain difficult across nearly all RUIs.
Their success depends on fine corrections that are poorly resolved by the available observations: a 2\,cm T-shirt misalignment spans only about eight pixels in the head camera.
Together, these cases expose the need for feedback and control mechanisms that operate at appropriate temporal and spatial resolutions.

\subsection{The Verification Gap}
\label{sec:verification}

\newlength{\judgew}
\begin{wraptable}{R}{0.57\linewidth}
  \vspace{-16pt}
  \caption{\textbf{Completion-judgment diagnostics (\%).}
  Online (30 tasks, one test seed): claim rate over episodes, false claims among completion claims, and completion claims among failed episodes.
  Post-hoc (23 tasks): failures detected in final frames by a fresh judge and claims made without rechecking.}
  \label{tab:judgment}
  \centering
  \footnotesize
  \setlength{\tabcolsep}{2pt}
  \setlength{\judgew}{33pt}
  \begin{NiceTabular*}{\linewidth}{@{\extracolsep{\fill}}l*{5}{w{c}{\judgew}}@{}}[cell-space-limits=1.3pt]
    \CodeBefore
      \rectanglecolor{slRed!14}{3-4}{4-4}
    \Body
    \toprule
    \Block[l]{2-1}{Agent}
      & \Block{1-3}{Online judgment} & &
      & \Block{1-2}{Post-hoc analysis} & \\
    \cmidrule(lr){2-4} \cmidrule(lr){5-6}
    & \makecell{Claim\\rate}
    & \makecell{False\\claims}
    & \makecell{Failed\\claimed}
    & \makecell{Failure\\visible}
    & \makecell{No\\recheck} \\
    \midrule
    Claude Code
      & 94.6 & 26.0 & \textbf{82.1} & 70.2 & 69.7 \\
    Codex
      & 91.8 & 21.8 & \textbf{70.9} & 80.9 & 92.1 \\
    \bottomrule
  \end{NiceTabular*}
  \vspace{-20pt}
\end{wraptable}

\paragraph{Finding 4: Observing failure does not guarantee recognizing it}
Beyond executing actions, an agent must determine whether the task has actually succeeded.
Yet 82.1\% and 70.9\% of failed episodes end in a completion claim for Claude Code and Codex, respectively (\cref{tab:judgment}).
We call this failure mode \textbf{illusory completion}.

Crucially, these failures are often visually detectable (\cref{fig:app-claims}).
In an offline analysis of 23 tasks (\cref{app:judge}), a fresh judge using only the final frames the agent had already received recognizes 70.2\% of Claude Code's false claims and 80.9\% of Codex's.
Most of these claims are made without another observation (69.7\% and 92.1\%, respectively).
For Codex, the offline judge uses the same underlying model in a fresh context, suggesting that the problem is not simply a lack of visual evidence.
Rather, \textbf{access to evidence does not \mbox{guarantee evidence-grounded completion judgments}}.

\paragraph{Action-level checks are insufficient for task-level verification}
We introduce post-condition checks for grasping and placing in the Skill and Program-Skill interfaces.
These checks use only observable robot states and camera measurements, achieving 88.6\% agreement with 149 human-verified labels (grasps 97.3\%, placements 80.0\%) and 94.5\% agreement with 524 simulator records.

Despite this agreement, the checks provide limited improvements (\cref{tab:ablation}).
Reporting failed actions lowers the share of failed episodes that end in a completion claim from 87.5\% to 81.2\%, and automatic retries lower it to 75.0\% without improving task success.
Only 2 of 32 corrections recover a failed grasp.
An agent may successfully execute individual actions while still failing the overall task.
This exposes a fundamental distinction: \textbf{action-level feedback cannot substitute for task-level verification}.

\section{Conclusion}
\label{sec:conclusion}

What a robot agent can accomplish depends not only on its model, but also on the interface through which it acts.
We introduced \RUIBench and \RUIKit to systematically study this relationship across 30 agentic manipulation tasks and 13 interfaces.
Our findings reveal that effective RUIs \textbf{reduce unnecessary decisions while preserving essential feedback}, with their benefits shaped by agent behavior and physical uncertainty.
Such interfaces are also easier for small models to learn.
We further uncover a \textbf{verification gap}, where agents frequently declare success despite observable failures.
Together, these results establish RUI design as a fundamental problem of allocating decisions, execution, and feedback between agents and robots.
\textbf{The key is not simply what actions an agent can perform, but when to decide, when to correct, and how to verify success.}

\paragraph{Limitations and future work}
Extending the real-robot study beyond five tasks on one arm would test how widely the observed interface shifts hold.
Contact-rich and time-critical tasks remain difficult across nearly all RUIs, calling for feedback and control at finer temporal and spatial resolutions, such as reactions inside programs or lower-level controllers.
Closing the verification gap further requires task-level verifiers that check completion against observations rather than individual actions.

\bibliographystyle{plainnat}
\bibliography{A-reference}

\clearpage
\appendix
\crefalias{section}{appendix}
\crefalias{subsection}{subappendix}

\section{Benchmark Details}
\label{app:benchmark}

\begin{table}[!h]
  \caption{\textbf{The 30 tasks of \RUIBench.}
  Source: adapted from a psychology paradigm (Psych.) or drawn from everyday games, puzzles, school experiments, and chores (Everyday).
  Tasks marked $^\dagger$ use two ARX~X5 arms, and the others use one Franka Research~3.
  Limit: time limit in simulated minutes, the turn budget times 8\,s, capped at 30\,min.
  Each task provides 15 development and 20 test missions.}
  \label{tab:app-tasks}
  \centering
  \footnotesize
  \renewcommand{\arraystretch}{1.0}
  \setlength{\tabcolsep}{3pt}
  \begin{NiceTabular*}{\textwidth}{@{\extracolsep{\fill}}w{l}{20pt}w{l}{63pt}w{l}{51pt}p{260pt}w{c}{21pt}@{}}[cell-space-limits=0.2pt]
    \toprule
    ID & Task & Source & Success requires & Limit \\
    \midrule
    \Block[l]{1-5}{\textbf{World knowledge}\enspace\emph{reasoning from prior knowledge}} \\
    WK1 & Tower of Hanoi & Psych.~\cite{simon1975functional} & All rings on the right peg, largest at the bottom, with every move legal. & 16.0 \\
    WK2 & River Crossing & Everyday & Wolf, goat, and bucket ferried across without leaving a forbidden pair. & 16.0 \\
    WK3 & Domino Run & Everyday & Dominoes placed so that a released pendulum topples them onto a bell. & 8.0 \\
    WK4 & Set the Table$^\dagger$ & Everyday & Fork, knife, and spoon placed around the plate as table manners require. & 12.0 \\
    WK5 & Balance Beam & Psych.~\cite{siegler1976three} & The beam leveled by placing the given weights in its pans. & 8.0 \\
    WK6 & Tic-Tac-Toe & Psych.~\cite{crowley1993flexible} & A game in progress won with three Xs in a row. & 5.3 \\
    \midrule
    \Block[l]{1-5}{\textbf{In-context learning}\enspace\emph{using previously shown visual examples}} \\
    CL1 & Play It Back$^\dagger$ & Everyday & A tune shown in a video replayed on a xylophone. & 5.3 \\
    CL2 & Dolls' Names & Psych.~\cite{carey1978acquiring} & The doll named in a video put to bed, with no other doll moved. & 4.0 \\
    CL3 & Build the House & Everyday & Blocks arranged as drawn on a card, with the structure standing still. & 8.0 \\
    CL4 & Crow and Pitcher & Psych.~\cite{bird2009rooks,cheke2012children} & A floating ball raised with stones and moved to a dish, as videos show. & 14.7 \\
    CL5 & Tangram & Everyday & All seven pieces placed on a silhouette, as in a finished example. & 12.0 \\
    CL6 & Knight's Move & Everyday & The black king captured by knight moves under a newly stated rule. & 8.0 \\
    \midrule
    \Block[l]{1-5}{\textbf{Discovery}\enspace\emph{physical exploration}} \\
    DS1 & Whose Present? & Everyday & Tom's present, found by reading the gift tags, placed on the red mat. & 8.0 \\
    DS2 & Egg Hunt & Everyday & An egg hidden under one of three pots found and put in the basket. & 16.0 \\
    DS3 & Treasure Hunt & Everyday & A coin under one of four shells found with a detector, lifting one shell. & 9.6 \\
    DS4 & Fake Coin & Everyday & The lighter of three coins found with a balance and put in the dish. & 10.1 \\
    DS5 & Find the Acid & Everyday & The acid among three identical beakers identified with pH strips. & 10.1 \\
    DS6 & Flame Test & Everyday & A green flame produced by testing splints soaked in unlabeled salts. & 11.7 \\
    \midrule
    \Block[l]{1-5}{\textbf{Dynamic interaction}\enspace\emph{responding to time-sensitive events}} \\
    DI1 & Rocket Launch$^\dagger$ & Psych.~\cite{lucas2014children} & A rocket launched by finding out how its two buttons must be pressed. & 4.0 \\
    DI2 & Load the Train & Everyday & Two toys loaded into the matching wagons of a moving train. & 10.7 \\
    DI3 & Traffic Light & Psych.~\cite{tominey2011red} & Three ducklings carried across the road only while the light is green. & 30.0 \\
    DI4 & Simon Says & Psych.~\cite{strommen1973verbal} & Every ``Simon says'' call carried out in time, and no other call. & 28.3 \\
    DI5 & Whack-a-Mole$^\dagger$ & Everyday & All moles but one hit before they retract. & 30.0 \\
    DI6 & Dessert Covers & Psych.~\cite{call2001object} & A tart found under one of three turned covers, lifting only one cover. & 15.5 \\
    \midrule
    \Block[l]{1-5}{\textbf{Contact-rich manipulation}\enspace\emph{precise adjustments during contact}} \\
    CR1 & Shape Sorter & Psych.~\cite{ornkloo2007fitting} & Four blocks dropped through the holes of their shapes. & 16.0 \\
    CR2 & Screw the Bulb & Everyday & A bulb screwed into a socket until the lamp lights. & 8.0 \\
    CR3 & Curling & Everyday & A stone pushed, never grasped, to rest inside the house. & 8.0 \\
    CR4 & Graphics Card & Everyday & A graphics card stood up and seated in a motherboard slot. & 12.0 \\
    CR5 & Snap the Bricks & Psych.~\cite{verdine2014deconstructing} & Two toy bricks pressed fully and squarely onto a tower. & 9.3 \\
    CR6 & Fold the T-Shirt$^\dagger$ & Everyday & A T-shirt folded from its bottom edge to its shoulders. & 8.0 \\
    \bottomrule
  \end{NiceTabular*}
\end{table}

\paragraph{Tasks}
\Cref{tab:app-tasks} lists the 30 tasks with their sources, success criteria, and time limits.
The 11 psychology tasks adapt classic paradigms of problem solving, word learning, causal learning, tool use, self-regulation, object permanence, and spatial cognition.
Success criteria combine the goal state with the rules of the task.
For example, the Tower of Hanoi also requires every move to be legal, and the Treasure Hunt allows the agent to lift only one shell.

\paragraph{Missions and splits}
A mission is a task instance generated from a fixed seed.
The seed sets the object poses and hidden variables, such as the pot that hides the egg.
Each task has 15 development missions (seeds 0 to 14) and 20 test missions (seeds 15 to 34).
\Cref{tab:main} uses the first five test missions of every task.
\Cref{tab:ablation,tab:judgment} and the simulation column of \cref{tab:real} use the first.

\paragraph{Limits and scoring}
Each task has a turn budget and a time limit on the episode clock, equal to the budget times 8\,s and capped at 30\,min.
The scorer of each task decides success at the end of the episode from the recorded object states and can recompute it from the logs.
Success requires the goal state, no violation of the task rules, and no alarm from a physics audit.
The audit flags numerical anomalies, articulated objects beyond their joint limits, \mbox{objects faster than 6\,m/s, and objects jumping over 3\,cm without contact.}

\paragraph{Negative controls}
The scripted reference solves all 70 control missions, the first five test missions of 14 tasks.
Each control runs it without one kind of information.
Without demonstrations, it fails on all 20 missions of CL1, CL2, CL4, and CL5; CL3 draws its goal on a card in the scene, and CL6 states its rule in text.
Planning from the first frame alone, it succeeds on 4 of 35 missions of discovery and dynamic tasks (DS1 to DS5, DI5, and DI6), each time by guessing.
Ignoring mid-episode changes, it succeeds on 2 of 15 missions of tasks whose target or instruction changes (CL6, DI2, and DI4).
Thus, \textbf{the tested tasks depend on the information that defines their suite}.

\paragraph{Observations}
SAM~3~\cite{carion2025sam} detects and segments objects in the rendered images, and sensor depth lifts them to 3D.
In simulation, each camera pose carries a calibration error of 3\,mm and 0.3\textdegree, fixed for a mission.
Depth carries a per-pixel error of 1\,mm plus 0.3\% of the distance, fixed for a viewpoint, so repeated looks do not average it out.
Positions therefore come from the sensors rather than the simulator, with smaller errors than on the real robot, where the two cameras can place an object up to 1.5\,cm apart (\cref{sec:real}).
After every reply that moves an arm, the harness \mbox{returns the latest images of all cameras with the result.}

\section{Interface Details}
\label{app:ruikit}

\begin{table}[t]
  \caption{\textbf{What each RUI exposes.}
  Action abstractions differ only in their action commands, and every abstraction except Step adds \tool{robot\_gripper}.
  Interaction forms decide how these commands are issued and when control returns to the agent, while the shared tools are identical in every RUI.}
  \label{tab:app-rui}
  \centering
  \footnotesize
  \renewcommand{\arraystretch}{1.0}
  \setlength{\tabcolsep}{3pt}
  \begin{NiceTabular*}{\textwidth}{@{\extracolsep{\fill}}w{l}{60pt}>{\raggedright\arraybackslash}p{164pt}>{\raggedright\arraybackslash}p{198pt}@{}}[cell-space-limits=0.2pt]
    \toprule
    Component & Tools & What one call specifies or returns \\
    \midrule
    \Block[l]{1-3}{\textbf{Action abstraction}\enspace\emph{varies across RUIs}} \\
    Joint & \tool{robot\_move\_joints}, \tool{robot\_solve\_ik} & Joint angles; the IK call returns the angles for a pose \\
    Delta & \tool{robot\_move\_delta} & A translation and rotation relative to the gripper \\
    Step & \tool{robot\_step} & A 0.25/1/4\,cm step or 2/5/15\textdegree{} turn, $n$ times, or a grasp \\
    Pose & \tool{robot\_move\_to} & An absolute position and orientation of the gripper \\
    Point & \tool{robot\_move\_to\_pixel} & A pixel lifted to 3D by depth, with an orientation \\
    Skill & \tool{skill\_\{pick, place, press,}\newline\tool{push, move, turn\}} & A whole operation on an object, a pixel, or a point \\
    \midrule
    \Block[l]{1-3}{\textbf{Interaction form}\enspace\emph{varies across RUIs}} \\
    Per-action & The abstraction's commands & At most one action per arm per reply \\
    Program & \tool{code\_run} & A Python program over the per-action tools \\
    RUI-Free & The tools of every abstraction in both forms & Any abstraction, in either form \\
    Action sequence & \tool{code\_run} over action commands only & Actions in order, with results at the end (\cref{tab:ablation} only) \\
    \midrule
    \Block[l]{1-3}{\textbf{Shared tools}\enspace\emph{identical in every RUI}} \\
    Camera & \tool{camera\_\{look, zoom\}} & Images of the fixed, head, or wrist cameras \\
    Vision & \tool{vision\_\{detect, segment, lift,}\newline\tool{pose, point, marks, track\}} & Detection, masks, 3D pose, pointing, marks, tracking \\
    Robot & \tool{robot\_\{state, wait, stop\}} & Joint and gripper readings, waiting, stopping \\
    Memory & \tool{memory\_\{note, recall, frames, scene\}} & Notes, earlier images, and a record of the scene \\
    Plan & \tool{plan\_\{grasps, motion, ik\}} & Grasps, collision-free paths, reachability of a pose \\
    Task & \tool{task\_\{info, demo, events,}\newline\tool{done, give\_up\}} & Instructions, demonstrations, events, episode end \\
    \bottomrule
  \end{NiceTabular*}
\end{table}

\paragraph{Controlled interface design}
RUIs differ only in what one action specifies and in when control returns to the agent (\cref{tab:app-rui}).
Every RUI offers the same 24 tools for perception, memory, planning, and the task, so the agent has the same means to perceive and plan under every RUI.
A program may call every per-action tool of its abstraction except \tool{task\_give\_up}.
The action-sequence variant of \cref{tab:ablation} runs a program that may only act, wait, and stop, and it returns the results of its actions when the program ends.
Its agent thus observes the scene and makes \mbox{decisions only between programs, as in the per-action form.}

\section{Experimental Details}
\label{app:experiments}

\subsection{Small Model}
\label{app:qwen}

\paragraph{Training data}
For each of the 12 fixed RUIs, the training set holds 300 trajectories from seeds disjoint from the test seeds.
Of these, 90 are successful Qwen3.8-Max episodes and 210 come from scripted policies.
Qwen3.8-Max runs in the same minimal loop and RUI as the small model, and we keep only the episodes that the benchmark scorer marks successful, without their reasoning.
The scripted policies plan with privileged state but act only through the commands of the RUI, with the same tools, clock, and budgets as at test time, so every trajectory is a valid episode of that RUI.
About half of the scripted episodes add an error to the target pose of the hand in 30\% to 60\% of the replies that move an arm, with standard deviations of up to 8\,mm in position and 4\textdegree{} in yaw, whatever the abstraction.
The policy then replans from the resulting state.
Replies whose execution carries such an error stay in the history of later steps but are not training targets, so the \mbox{model learns the corrections without learning to make the errors.}

\paragraph{Fine-tuning and inference}
Each RUI has its own LoRA adapter~\cite{hu2022lora} of rank 64 ($\alpha=128$) on all linear layers of the language model, while the vision encoder stays frozen.
All 12 adapters share the data recipe, the number of trajectories, and the training schedule.
Training runs for 10 epochs with a learning rate of $10^{-4}$, cosine decay, and 32 requests per batch.
Each training example is one request exactly as \RUIKit sends it, with the system prompt, the tools, the text history, and the latest six images, and the loss covers only the reply.
At test time, the model runs in the minimal loop without thinking, with the same tools, limits, and 8\,s clock as the frontier agents, on the 150 test episodes of each RUI.

\paragraph{Episode length and cost}
Abstractions that name a target shorten the small model's episodes.
In the per-action form, Point and Skill episodes take a median of 29 and 23 replies, against 53 for Joint and 36 for Delta.
Programs cut Joint, the longest, to 37.5 replies, in line with their gain on Joint alone (\cref{sec:sim}).
Train cost charges \$2.68 per H200 GPU-hour and \$4.06 per B200 GPU-hour.
Test cost charges one dedicated inference GPU that keeps the model loaded, as a deployed robot would, for the wall-clock duration of each test episode, including the time spent executing actions.

\subsection{Programs and Feedback}
\label{app:programs}

\begin{table}[t]
  \centering
    \caption{\textbf{Programs written by the two agents.}
  Six program RUIs on the first test mission (180 episodes per agent).
  Actions: mean actions per program, where a program that only observes counts zero; ${\geq}5$ actions, Loops, and Observe: share of programs with at least five actions, with loops, and with a perception call after an action (\%); Lines: median lines per program.
  Paired wins: pairs of task and abstraction in which only one form succeeds, won by the program and by the per-action form.}
  \label{tab:app-programs}
  \footnotesize
  \setlength{\tabcolsep}{2pt}
  \begin{NiceTabular*}{\textwidth}{@{\extracolsep{\fill}}w{l}{56pt}*{6}{w{c}{50pt}}@{}}[cell-space-limits=1pt]
    \toprule
    Agent & Actions & ${\geq}5$ actions & Loops & Observe & Lines & Paired wins \\
    \midrule
    Claude Code & 3.3 & 21 & 58 & 54 & 7 & 32 : 8 \\
    Codex & 0.9 & 0 & 10 & 8 & 1 & 20 : 14 \\
    \bottomrule
  \end{NiceTabular*}

  \vspace{12pt}
    \caption{\textbf{Interaction-form ablation of \cref{tab:ablation} by suite and abstraction.}
  Success (\%) of Claude Code on the first test mission, with 18 episodes per suite and 30 per abstraction.
  Suites: world knowledge (WK), in-context learning (CL), discovery (DS), dynamic interaction (DI), and contact-rich manipulation (CR).}
  \label{tab:app-suites}
  \footnotesize
  \setlength{\tabcolsep}{2pt}
  \begin{NiceTabular*}{\textwidth}{@{\extracolsep{\fill}}w{l}{62pt}*{9}{w{c}{30pt}}@{}}[cell-space-limits=1pt]
    \CodeBefore
      \rowcolor{slBlue!6}{3}
      \rowcolor{slGreen!4}{4}
      \rowcolor{slGreen!8}{5}
    \Body
    \toprule
    \Block[l]{2-1}{Form} & \Block{1-5}{Suite} & & & & & \Block{1-3}{Abstraction} & & & \Block{2-1}{All} \\
    \cmidrule(lr){2-6} \cmidrule(lr){7-9}
    & WK & CL & DS & DI & CR & Delta & Step & Point & \\
    \midrule
    Per-action & 83.3 & 33.3 & 66.7 & 50.0 & 55.6 & 66.7 & 46.7 & 60.0 & 57.8 \\
    Action sequence & 83.3 & 66.7 & 72.2 & 55.6 & 50.0 & 70.0 & 60.0 & 66.7 & 65.6 \\
    Program & 100.0 & 66.7 & 88.9 & 72.2 & 50.0 & 80.0 & 70.0 & 76.7 & 75.6 \\
    \bottomrule
  \end{NiceTabular*}

\end{table}

\paragraph{Programs help when agents compose actions}
Programs raise Claude Code's success by 11.0 points but Codex's by only 2.7 points (\cref{tab:main}).
\Cref{tab:app-programs} explains the difference.
Claude Code's programs contain 3.3 actions on average, and 54\% of them observe the world after acting.
Codex's programs contain 0.9 actions on average, and its median program is a single line.
On the same tasks and abstractions, Claude Code's programs win 32 of the 40 pairs in which only one form succeeds, while Codex's win 20 of 34.
Thus, \textbf{the program form pays off \mbox{when the agent uses it to compose actions with feedback}}.

\paragraph{Composition and feedback by suite}
\Cref{tab:app-suites} breaks the interaction-form ablation of \cref{tab:ablation} down by suite and abstraction.
For each of the three abstractions, success rises from per-action execution to action sequences and again to programs, and the suites follow the pattern that \cref{sec:ablation} describes.

\subsection{Real Robot}
\label{app:real}

\begin{figure}[t]
  \centering
  \small
  \newcommand{\rolloutrow}[2]{
    \parbox{\textwidth}{\raggedright\textbf{#1}}\\[2pt]
    \includegraphics[width=0.246\textwidth]{figures/rollouts/#21.jpg}\hfill
    \includegraphics[width=0.246\textwidth]{figures/rollouts/#22.jpg}\hfill
    \includegraphics[width=0.246\textwidth]{figures/rollouts/#23.jpg}\hfill
    \includegraphics[width=0.246\textwidth]{figures/rollouts/#24.jpg}\\[4pt]}
  \rolloutrow{(a) Domino Run (WK3), Claude Code with Delta: success after 27 replies}{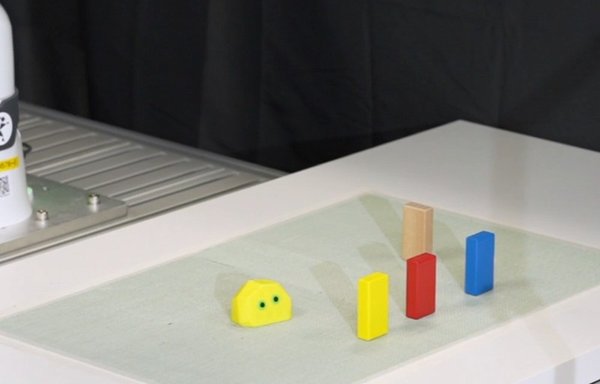}
  \rolloutrow{(b) Domino Run (WK3), Claude Code with Program-Delta: failure after 5 replies}{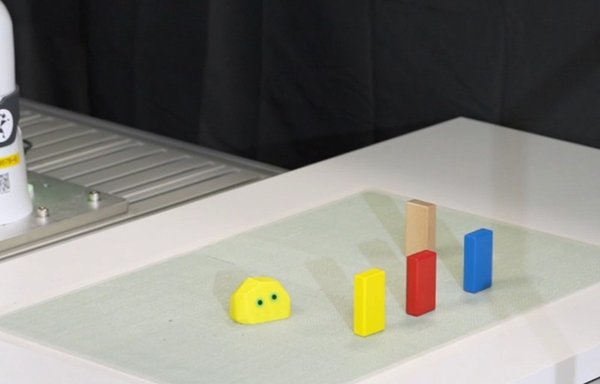}
  \rolloutrow{(c) Build the House (CL3), Codex with Skill: success after 31 replies}{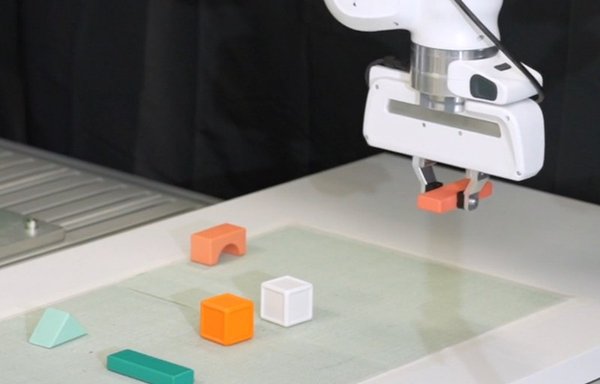}
  \rolloutrow{(d) Shape Sorter (CR1), Claude Code with Skill: success after 64 replies}{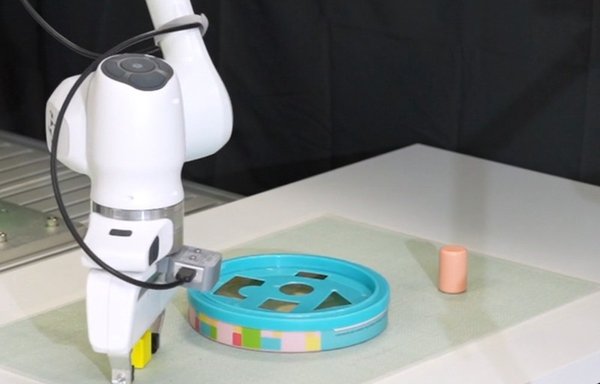}
  \vspace{-4pt}
  \caption{\textbf{Real-robot rollouts on the first layout.}
  Each row shows four frames of one episode from the third-person camera.
  On the same task and layout, Claude Code topples the whole chain with per-action Delta (a), but only the blue domino falls, sideways, after its program (b).
  Per-action episodes also recover from a beam that lands beside its supports (c) and a cube that rests tilted on the lid (d).}
  \label{fig:app-real}
\end{figure}

\paragraph{Setup}
The real station uses the Franka Research~3 arm of \cref{sec:real} with a Franka Hand, a fixed RealSense D435 across the table, and a wrist-mounted D405.
The harness, RUIs, tools, and perception code run unchanged on a real-robot backend.
The backend sends setpoints at 50\,Hz to impedance controllers, which follow minimum-jerk paths under speed limits.
Before each motion, the backend previews the path on a kinematic model and checks the workspace, hand tilt, and joint limits.
A motion that cannot be completed stops at its last reachable point and returns \emph{blocked}, as in simulation.
During the motion, the arm stops when tracking error, joint positions, or joint speeds exceed their limits.
Perception adds no synthetic noise to the real sensors.
A ChArUco board on the table calibrates both cameras, and time follows the wall clock, including the agent's reasoning.
A separate camera, which the agent never sees, records each episode for the figures and videos.
An operator supervises every episode for safety and can pause the arm or stop the episode, which then counts as a failure.

\paragraph{Tasks and scoring}
Each real task keeps the goal of its simulated counterpart with physical objects (\cref{fig:tasks}); for example, the dominoes must topple onto a toy rather than ring a bell.
Each task runs on five fixed layouts, the same for every RUI and agent, with the turn budgets and time limits of simulation.
Because the objects differ, Finding~3 compares interfaces within each world.
After each episode, the operator judges success by the written criterion of the real task and rates each success from 1 to 5 in half-point steps.
We count failures as 0 and rescale the ratings to 100.
The ratings preserve the orderings of \cref{tab:real}, with per-action execution ahead of programs \mbox{for both Claude Code (30.1 vs.\ 27.7) and Codex (44.1 vs.\ 41.5).}

\paragraph{Rollouts}
\Cref{fig:app-real} shows four real-robot episodes on the first layout.
On the Domino Run, Claude Code with per-action Delta builds a chain that topples into the stone toy (a).
With Program-Delta, it places the dominoes and ends the episode after 5 replies instead of 27, but the chain breaks at the first domino (b).
Successful episodes often place an object, inspect the result, and correct it, as the beam in (c) and the cube in (d) show.
Seventeen more \mbox{real-robot episodes and 29 simulated ones follow in \cref{app:gallery}.}

\subsection{Completion Judgment}
\label{app:judge}

\begin{figure}[t]
  \centering
  \small
  \newcommand{\claimpanel}[4]{
    \begin{minipage}[t]{0.485\textwidth}
      \includegraphics[width=\linewidth]{#1}\\[3pt]
      \raggedright
      \textbf{#2}\\
      \textbf{Agent:} #3\\
      \textbf{Judge:} #4
    \end{minipage}}
  \claimpanel{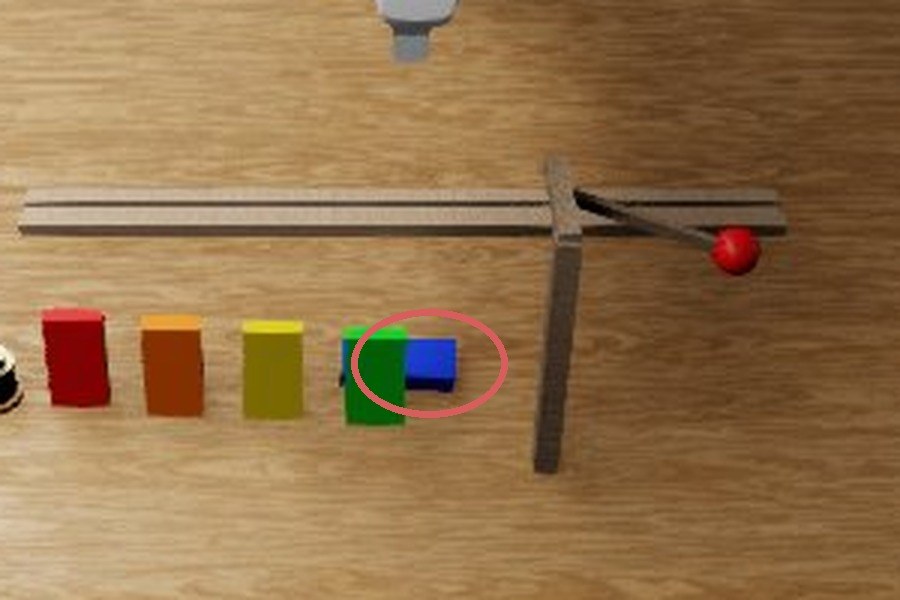}{(a) Domino Run (WK3), Codex with Joint}
    {``Placed four upright dominoes in a chain toward the bell and withdrew the arm; one blue domino fell during handling.''}
    {``The blue domino nearest the pendulum is lying flat, leaving no continuous upright chain for the pendulum to topple toward the bell.''}\hfill
  \claimpanel{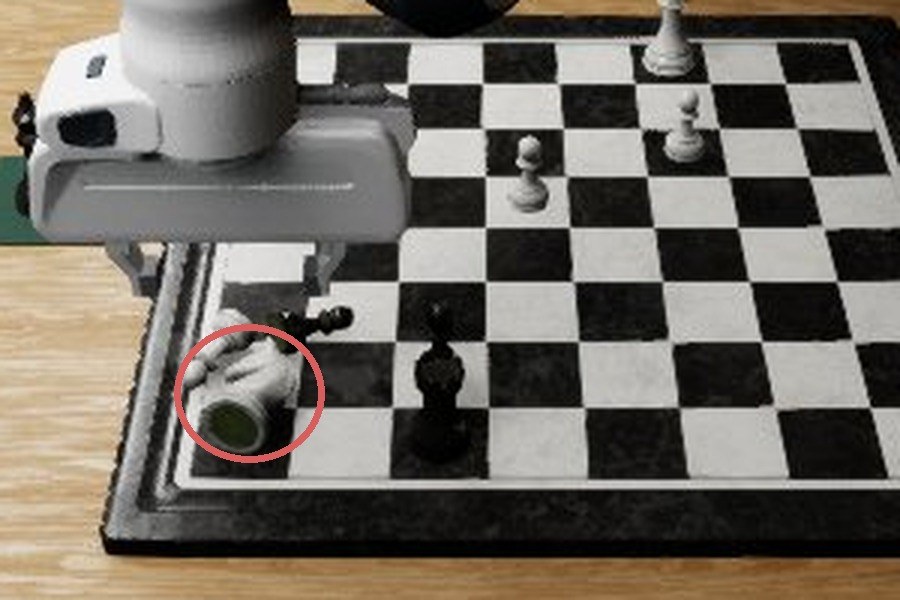}{(b) Knight's Move (CL6), Codex with RUI-Free}
    {``Removed the black king and placed the white knight on its square; the knight is lying sideways after recovery, and a failed wrist rotation disturbed two nearby pawns.''}
    {``\ldots the white knight lies near the board's edge rather than occupying the king's square.''}
  \caption{\textbf{Illusory completion in simulation.}
  Each column crops the last camera image returned to the agent before it called \tool{task\_done}, and the red circle marks the failure that the judge names.
  Codex notes the fallen piece yet reports the goal as reached, whereas the same model, given only these images in a fresh context, recognizes the failure: observing a failure does not guarantee recognizing it.}
  \label{fig:app-claims}
\end{figure}

\paragraph{Post-hoc judge}
The judge is GPT-6 Astra in a fresh context.
It receives the instruction, the hint, and the last images of every camera that the harness returned to the agent before it called \tool{task\_done}.
It decides success or failure from these images alone and states its reason.
The post-hoc set of \cref{tab:judgment} keeps the 23 tasks on which the judge can recognize success from these images.
On six tasks, the outcome is hard to see in a final frame, such as a tune that was played (CL1) or an acid that was identified (DS5), and the judge rejects many true completions.
We leave out these tasks (CL1, CR2, DI4, DI5, DS5, and WK2) so that judge errors do not count as visible failures, together with CR6, on which neither agent succeeds on the first test mission.
On the 23 tasks, the judge confirms 95.4\% of Claude Code's true completions and 98.7\% of Codex's.
Its detection of false claims therefore reflects failures that the final frames show, as in the two Codex episodes shown in \cref{fig:app-claims}.

\section{Rollout Gallery}
\label{app:gallery}

\newcommand{\realrow}[4]{
  \parbox{\textwidth}{\raggedright\textbf{(#1) #2:}\\#3}\\[2pt]
  \includegraphics[width=0.235\textwidth]{figures/gallery/real/#4-1.jpg}\hfill
  \includegraphics[width=0.235\textwidth]{figures/gallery/real/#4-2.jpg}\hfill
  \includegraphics[width=0.235\textwidth]{figures/gallery/real/#4-3.jpg}\hfill
  \includegraphics[width=0.235\textwidth]{figures/gallery/real/#4-4.jpg}\\[5pt]}
\newcommand{\simrow}[2]{
  \parbox{\textwidth}{\raggedright\textbf{#1}}\\[2pt]
  \includegraphics[width=0.19\textwidth]{figures/gallery/sim/#2-1.jpg}\hfill
  \includegraphics[width=0.19\textwidth]{figures/gallery/sim/#2-2.jpg}\hfill
  \includegraphics[width=0.19\textwidth]{figures/gallery/sim/#2-3.jpg}\hfill
  \includegraphics[width=0.19\textwidth]{figures/gallery/sim/#2-4.jpg}\hfill
  \includegraphics[width=0.19\textwidth]{figures/gallery/sim/#2-5.jpg}\\[6pt]}

\begin{figure}[p]
  \centering
  \small
  \simrow{WK1, Tower of Hanoi: Claude Code with Program-Skill, success after 8 replies}{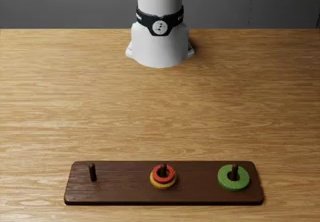}
  \simrow{WK2, River Crossing: Codex with Program-Skill, success after 23 replies}{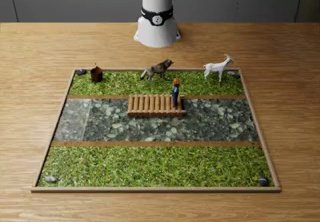}
  \simrow{WK3, Domino Run: Claude Code with Program-Pose, success after 11 replies}{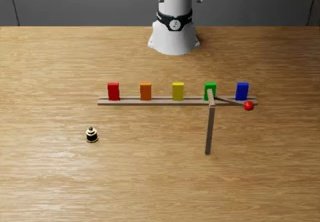}
  \simrow{WK4, Set the Table: Codex with Program-Joint, success after 25 replies}{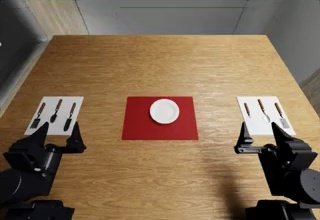}
  \simrow{WK5, Balance Beam: Claude Code with RUI-Free, success after 9 replies}{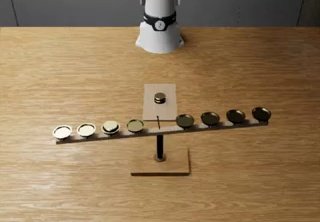}
  \simrow{WK6, Tic-Tac-Toe: Codex with Program-Point, success after 11 replies}{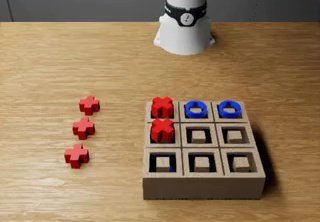}
  \vspace{-6pt}
  \caption{\textbf{Simulated rollouts: World Knowledge.}
  One successful episode per task on the first test mission, seen from the camera that the agent itself observes (the fixed camera, or the head camera for two-arm tasks).}
  \label{fig:app-gallery-wk}
\end{figure}

\begin{figure}[p]
  \centering
  \small
  \simrow{CL1, Play It Back: Claude Code with Program-Point, success after 10 replies}{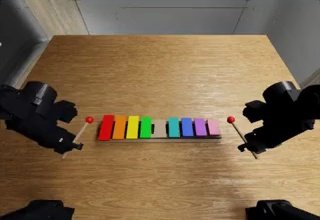}
  \simrow{CL2, Dolls' Names: Codex with Program-Delta, success after 12 replies}{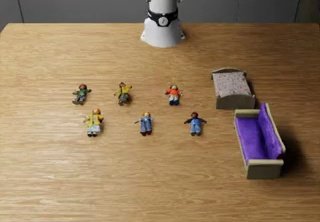}
  \simrow{CL3, Build the House: Claude Code with Pose, success after 26 replies}{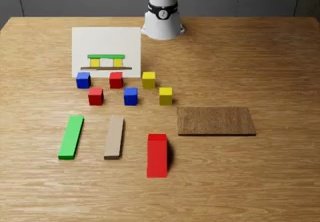}
  \simrow{CL4, Crow and Pitcher: Codex with Pose, success after 36 replies}{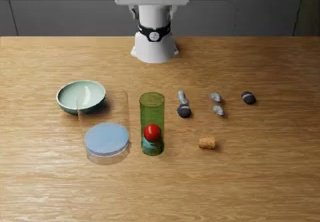}
  \simrow{CL5, Tangram: Claude Code with Program-Skill, success after 24 replies}{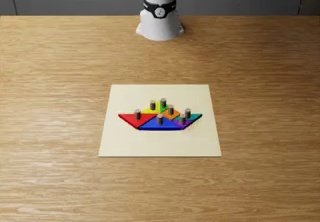}
  \simrow{CL6, Knight's Move: Codex with Program-Pose, success after 44 replies}{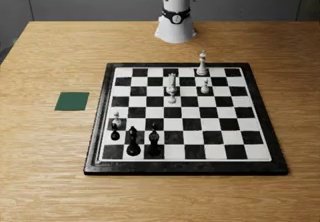}
  \vspace{-6pt}
  \caption{\textbf{Simulated rollouts: In-Context Learning.}
  One successful episode per task on the first test mission, seen from the camera that the agent itself observes (the fixed camera, or the head camera for two-arm tasks).}
  \label{fig:app-gallery-cl}
\end{figure}

\begin{figure}[p]
  \centering
  \small
  \simrow{DS1, Whose Present?: Claude Code with Delta, success after 13 replies}{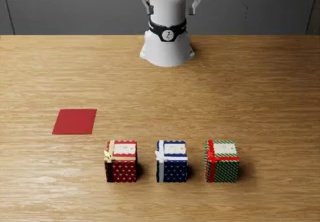}
  \simrow{DS2, Egg Hunt: Codex with Program-Step, success after 28 replies}{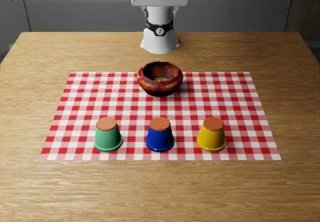}
  \simrow{DS3, Treasure Hunt: Claude Code with Program-Delta, success after 16 replies}{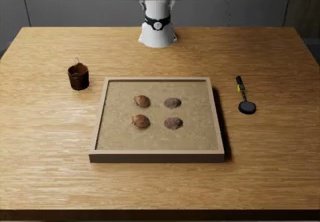}
  \simrow{DS4, Fake Coin: Codex with Step, success after 39 replies}{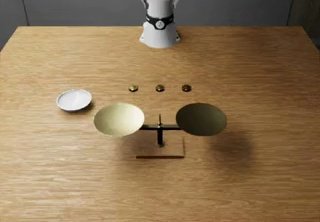}
  \simrow{DS5, Find the Acid: Claude Code with Program-Point, success after 32 replies}{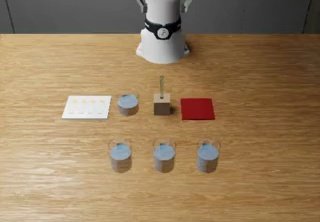}
  \simrow{DS6, Flame Test: Codex with RUI-Free, success after 24 replies}{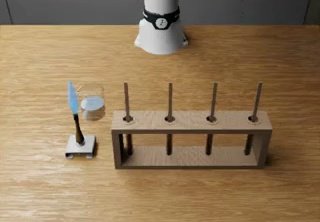}
  \vspace{-6pt}
  \caption{\textbf{Simulated rollouts: Discovery.}
  One successful episode per task on the first test mission, seen from the camera that the agent itself observes (the fixed camera, or the head camera for two-arm tasks).}
  \label{fig:app-gallery-ds}
\end{figure}

\begin{figure}[p]
  \centering
  \small
  \simrow{DI1, Rocket Launch: Claude Code with Skill, success after 5 replies}{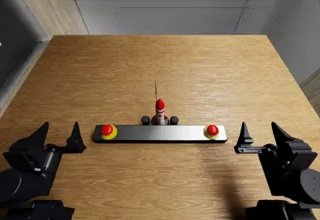}
  \simrow{DI2, Load the Train: Codex with Program-Point, success after 15 replies}{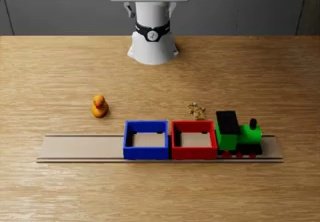}
  \simrow{DI3, Traffic Light: Claude Code with Program-Step, success after 12 replies}{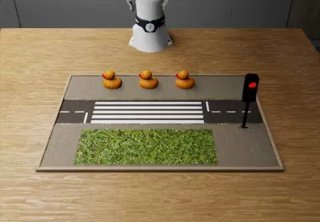}
  \simrow{DI4, Simon Says: Codex with Skill, success after 10 replies}{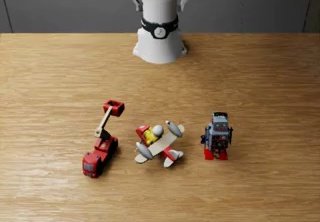}
  \simrow{DI5, Whack-a-Mole: Claude Code with Pose, success after 23 replies}{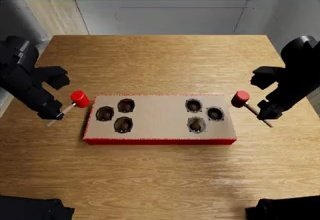}
  \simrow{DI6, Dessert Covers: Codex with Program-Joint, success after 27 replies}{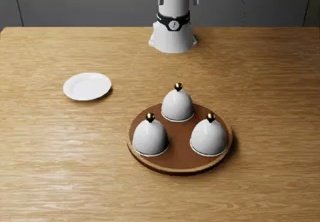}
  \vspace{-6pt}
  \caption{\textbf{Simulated rollouts: Dynamic Interaction.}
  One successful episode per task on the first test mission, seen from the camera that the agent itself observes (the fixed camera, or the head camera for two-arm tasks).}
  \label{fig:app-gallery-di}
\end{figure}

\begin{figure}[p]
  \centering
  \small
  \simrow{CR1, Shape Sorter: Claude Code with Joint, success after 53 replies}{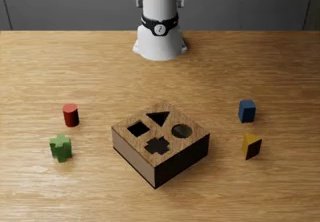}
  \simrow{CR2, Screw the Bulb: Codex with Joint, success after 21 replies}{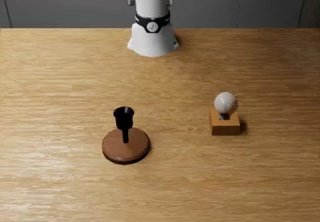}
  \simrow{CR3, Curling: Claude Code with Point, success after 11 replies}{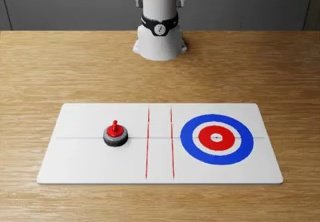}
  \simrow{CR4, Graphics Card: Codex with Program-Joint, success after 37 replies}{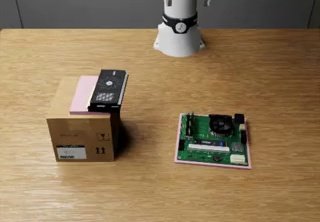}
  \simrow{CR5, Snap the Bricks: Claude Code with Step, success after 43 replies}{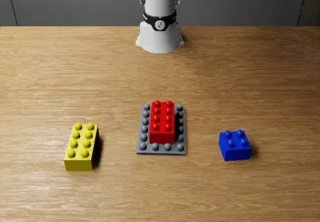}
  \vspace{-6pt}
  \caption{\textbf{Simulated rollouts: Contact-Rich Manipulation.}
  One successful episode per task on the first test mission, seen from the camera that the agent itself observes (the fixed camera, or the head camera for two-arm tasks). Fold the T-Shirt (CR6) has no successful episode on this mission.}
  \label{fig:app-gallery-cr}
\end{figure}

\begin{figure}[p]
  \centering
  \small
  \realrow{a}{Domino Run (WK3), Claude Code with Pose, success after 44 replies}{stands the dominoes up, then adjusts each one}{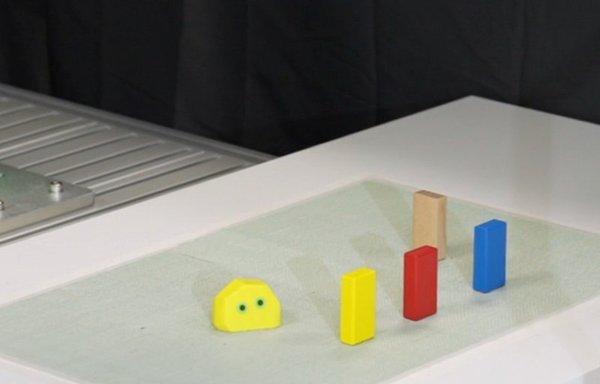}
  \realrow{b}{Domino Run (WK3), Codex with Pose, success after 53 replies}{straightens the row domino by domino}{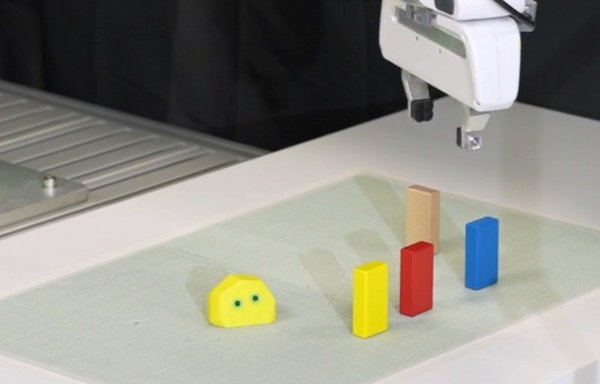}
  \realrow{c}{Domino Run (WK3), Codex with Delta, success after 41 replies}{repositions a domino after the first arrangement}{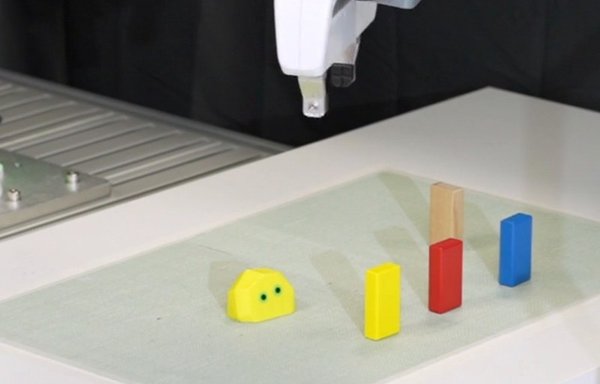}
  \realrow{d}{Build the House (CL3), Codex with Pose, success after 32 replies}{corrects the beam on its supports}{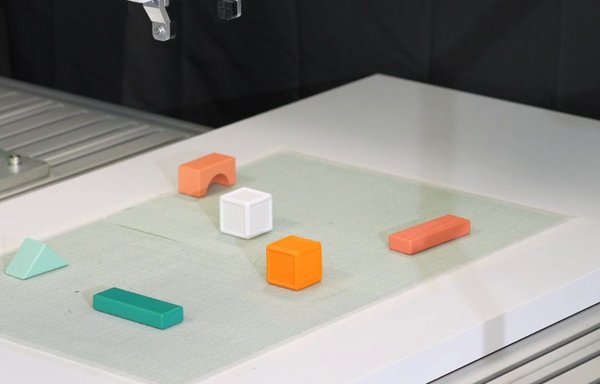}
  \realrow{e}{Build the House (CL3), Claude Code with RUI-Free, success after 13 replies}{the beam slides off, so it moves the cube closer and lays the beam again}{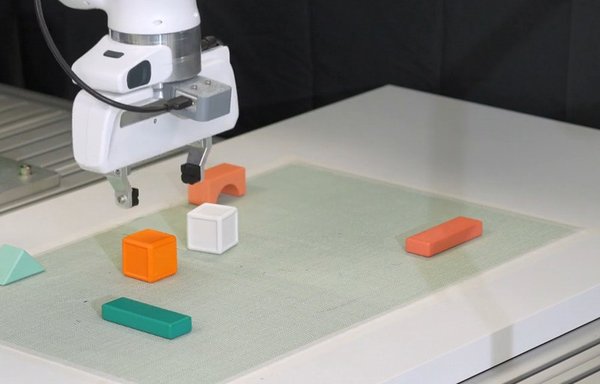}
  \realrow{f}{Build the House (CL3), Claude Code with Program-Pose, success after 10 replies}{the house collapses, and the program builds it again}{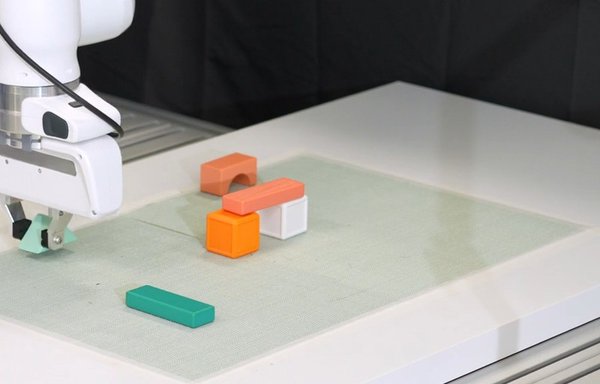}
  \vspace{-6pt}
  \caption{\textbf{Real-robot rollouts: correcting placements.}
  Each row shows four frames of one episode on the first layout, from the third-person camera, together with its task, agent, RUI, and number of replies. The agents inspect what they have placed and adjust it until the arrangement works.}
  \label{fig:app-gallery-real-a}
\end{figure}

\begin{figure}[p]
  \centering
  \small
  \realrow{a}{Shape Sorter (CR1), Codex with Delta, success after 46 replies}{the cube sticks at its hole and is pushed in}{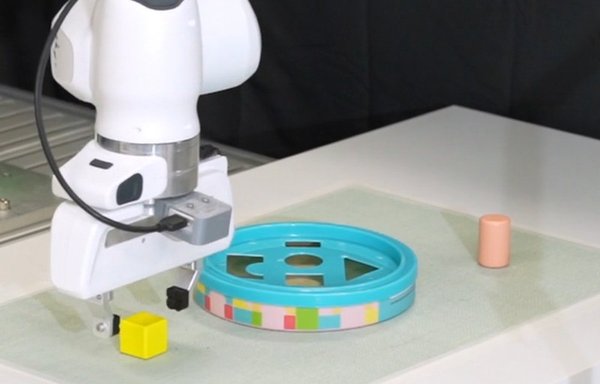}
  \realrow{b}{Shape Sorter (CR1), Claude Code with Program-Point, success after 45 replies}{presses the tilted cube into its hole}{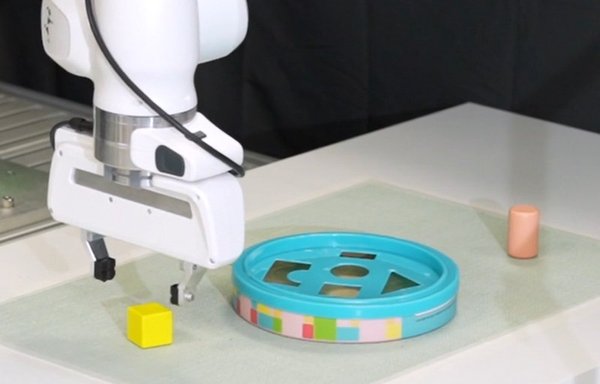}
  \realrow{c}{Shape Sorter (CR1), Claude Code with Pose, success after 89 replies}{presses, re-grasps, and nudges the cube into its hole}{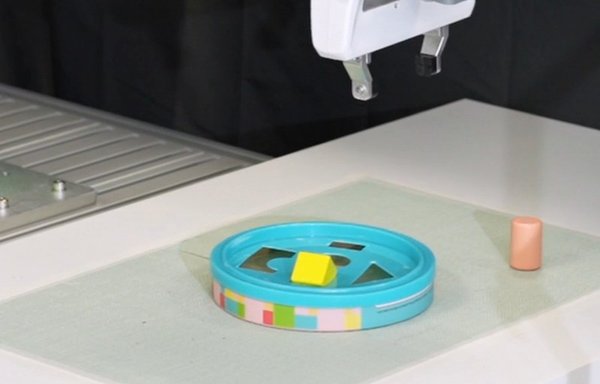}
  \realrow{d}{Build the House (CL3), Claude Code with Delta, success after 35 replies}{turns the hand by 90\textdegree{} to grip the beam}{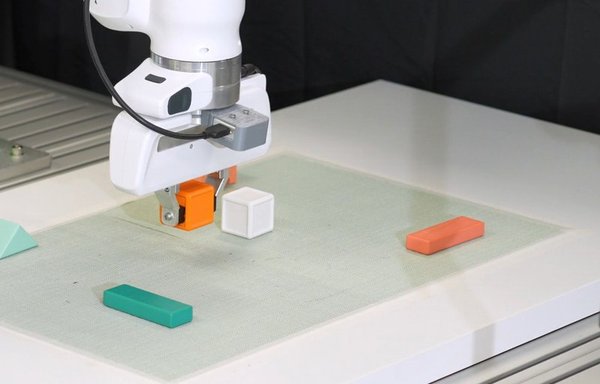}
  \realrow{e}{Egg Hunt (DS2), Claude Code with Step, success after 42 replies}{turns the hand to the faces of the small cube}{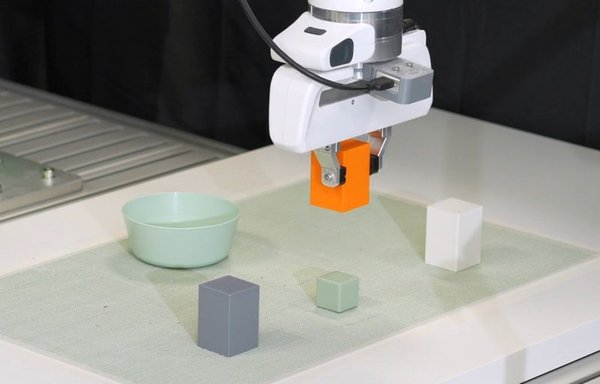}
  \realrow{f}{Egg Hunt (DS2), Claude Code with Point, success after 47 replies}{tilts the gripper to reach into the bowl}{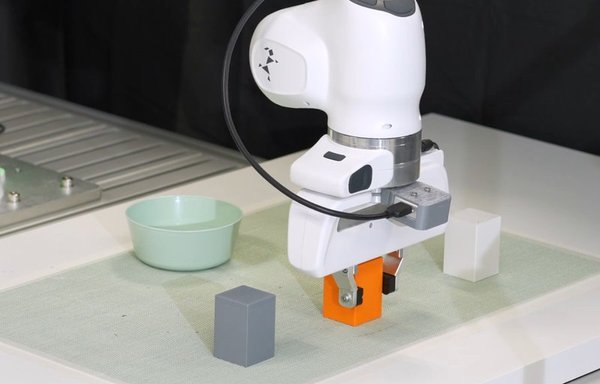}
  \vspace{-6pt}
  \caption{\textbf{Real-robot rollouts: contact and hand pose.}
  Each row shows four frames of one episode on the first layout, from the third-person camera, together with its task, agent, RUI, and number of replies. The agents push stuck or tilted pieces into their holes and turn or tilt the hand to suit the object and the target.}
  \label{fig:app-gallery-real-b}
\end{figure}

\begin{figure}[p]
  \centering
  \small
  \realrow{a}{Build the House (CL3), Claude Code with Program-Step, success after 8 replies}{builds the house with a few programs}{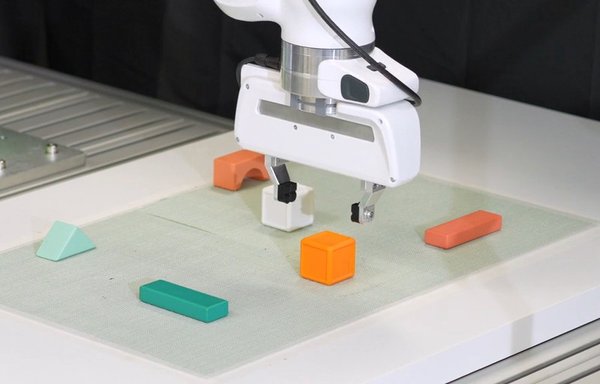}
  \realrow{b}{Build the House (CL3), Claude Code with Program-Delta, success after 8 replies}{places the base, the beam, and the roof}{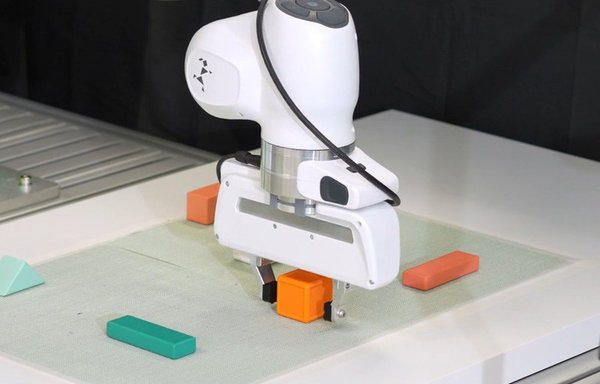}
  \realrow{c}{Domino Run (WK3), Codex with Program-Pose, success after 45 replies}{corrects the dominoes after the first arrangement}{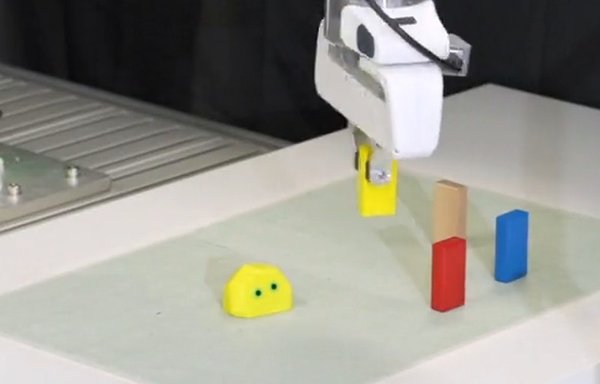}
  \realrow{d}{Shape Sorter (CR1), Codex with Program-Point, success after 37 replies}{inserts the cube and then the cylinder}{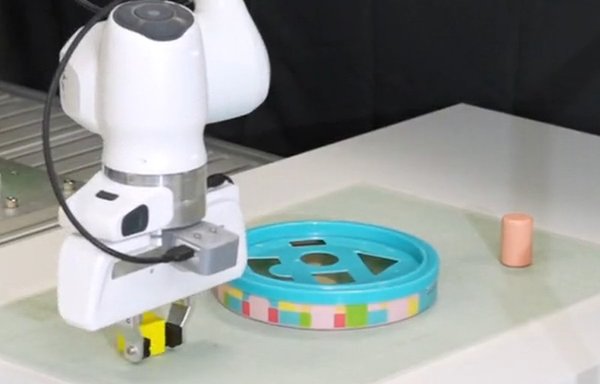}
  \realrow{e}{Egg Hunt (DS2), Codex with Pose, success after 22 replies}{moves the orange block, finds the small block under it, and puts it in the bowl}{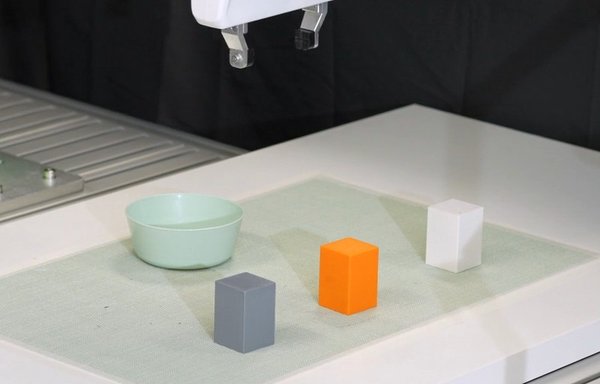}
  \vspace{-6pt}
  \caption{\textbf{Real-robot rollouts: programs and search.}
  Each row shows four frames of one episode on the first layout, from the third-person camera, together with its task, agent, RUI, and number of replies. Claude Code's programs compose several actions per reply (a, b), and a program episode can still correct its own work (c).}
  \label{fig:app-gallery-real-c}
\end{figure}

\Crefrange{fig:app-gallery-wk}{fig:app-gallery-cr} show one successful simulated episode for each of the 29 tasks solved on the first test mission, alternating between the two agents and covering all 13 RUIs.
\Crefrange{fig:app-gallery-real-a}{fig:app-gallery-real-c} show 17 more real-robot episodes on the first layout, grouped by what the agents do.
Each row names the task, the agent, the RUI, and the number of replies, and its frames run left to right in time.

\end{document}